\documentclass{article}

\usepackage[preprint]{neurips_2026}
\usepackage{listings}

\renewcommand{\acksection}{\section*{Acknowledgments}}

\usepackage[utf8]{inputenc}
\usepackage[T1]{fontenc}
\usepackage{hyperref}
\usepackage{url}
\usepackage{booktabs}
\usepackage{amsfonts}
\usepackage{amsmath}
\usepackage{amssymb}
\usepackage{nicefrac}
\usepackage{microtype}
\usepackage{xcolor}
\usepackage{graphicx}

\usepackage{multirow}
\usepackage{makecell}
\usepackage{array}

\usepackage{placeins}

\title{Visual-to-Executable Procedural Reconstruction of Buildings}

\author{%
Zichong Zhu$^{1}$,
Zhihao Liu$^{2}$,
Andrei Sharf$^{3}$,
Zhanglin Cheng$^{1}$\thanks{Corresponding author: \texttt{zl.cheng@siat.ac.cn}}\\
\\
$^{1}$Shenzhen VisuCA Key Lab, Shenzhen Institutes of Advanced Technology, Chinese Academy of Sciences\\
$^{2}$The University of Tokyo, $^{3}$Ben-Gurion University of the Negev
}

\begin{document}

\maketitle

\begin{abstract}
Procedural modeling offers compact, editable, and reusable representations for architectural content, but building such models still depends heavily on manually authored grammars. We study image-based procedural building reconstruction from in-the-wild images, with the goal of recovering an executable procedural description that can be instantiated as an editable 3D building asset. Because a single image provides only partial and view-dependent evidence, we focus on recovering the dominant visible facade(s) and attach the recovered facade program to a simple mass model for building-level instantiation. Our method decomposes this task into two structured prediction stages. First, a Global Structure Layout Tree recovers facade-scale split-and-repeat organization through a fine-tuned layout model with render-guided visual-symbolic reflection. Second, a Component Parameterization schema maps representative facade tiles to structured procedural attributes for local architectural component synthesis. These two representations are then deterministically compiled into executable Computer Generated Architecture (CGA) programs, avoiding the brittleness of direct low-level program generation. Experiments show that the proposed framework handles both regular facades and challenging in-the-wild images, while maintaining high structural consistency, procedural executability, and fine-grained architectural detail. The recovered programs further enable controllable and reusable 3D building asset generation.

\end{abstract}

\section{Introduction}

Procedural modeling is a compelling paradigm for architectural content creation. Buildings are naturally composed of repeated elements, hierarchical subdivisions, and parameterized components, making procedural descriptions particularly well-suited to their structure~\citep{wonka2003instant,muller2006procedural}. Compared with explicit geometry or purely neural representations~\citep{mildenhall2021nerf,kerbl20233d,poole2022dreamfusion}, procedural descriptions are compact, editable, and reusable, making them attractive for scalable and controllable 3D building asset generation. Yet, constructing such models still largely depends on manually authored grammars and domain expertise, which severely limits their practical scalability.

In this paper, we study image-based procedural building reconstruction from an in-the-wild image. Our goal is to recover an executable and editable procedural description that can be instantiated as a 3D building asset. Since a single image provides only partial and view-dependent evidence, we focus on the dominant visible facade(s) and treat the recovered program as a procedural interface for controllable building asset generation, while unseen geometry is instantiated using simple procedural completion rules.

Existing approaches only partially address this goal. Facade parsing methods recover semantic labels or dense structural cues, but do not produce editable or executable procedural programs. Classical inverse procedural modeling methods exploit architectural regularity, but typically assume rectified views, segmented layouts, or predefined grammar families. More recently, learning-based and LLM/VLM-based approaches suggest the possibility of image-to-program prediction, yet direct low-level program generation remains brittle because valid procedural programs are non-unique, tightly constrained by grammar validity, and hard to verify during prediction.

Our key observation is that the difficulty of this problem lies not only in generating code, but in bridging ambiguous visual evidence and execution-constrained procedural representations. The same building facade may admit multiple valid procedural decompositions with no single correct answer, while in-the-wild images further introduce perspective distortion, occlusion, and partial visibility, making it difficult to infer a clean orthogonal split-and-repeat scaffold directly from raw RGB evidence. We therefore avoid direct low-level image-to-code generation and instead cast the problem as verifiable image-to-structure and image-to-attributes prediction, followed by deterministic compilation into executable programs.

This formulation raises four coupled challenges. First, facade-scale organization is hierarchical and non-unique: the same facade may admit different split orders, repeat abstractions, or grouping strategies, making direct supervision on a single low-level program unstable. Second, in-the-wild images provide incomplete and view-distorted observations: perspective effects, occlusion, illumination variation, and partial visibility make it difficult to infer a clean orthogonal split-and-repeat scaffold from a non-rectified image. Third, local architectural elements such as windows, entrances, arches, sills, stairs, and hardware must be recovered as procedural parameters rather than merely detected as visual instances. Fourth, the recovered global scaffold and local components must be compositionally compatible and directly executable. A visually plausible output is insufficient if it violates grammar constraints or cannot be compiled.

To address these challenges, we propose a hierarchical neuro-symbolic framework with two structured intermediate representations. At the global level, we recover a Global Structure Layout Tree that captures facade-scale split-and-repeat organization and is iteratively refined through a render-guided visual-symbolic reflection process, making structural mismatches easier to identify and correct than direct symbolic comparison would allow. At the local level, we recover a Component Parameterization for representative repeated elements, mapping tile crops to structured attributes rather than raw procedural code. This design turns under-specified local geometry into a constrained and verifiable prediction problem.

The two levels are then assembled by deterministic compilers into executable CGA programs, producing a procedural building representation that is both editable and directly instantiable as a 3D asset. By separating global structural reasoning, local procedural grounding, and deterministic execution, our framework improves verifiability, robustness, and executability over direct low-level code generation.

Our contributions are threefold:
\begin{enumerate}
      \item We formulate in-the-wild image-based procedural building reconstruction as a two-level neuro-symbolic prediction problem and introduce a representation bridge consisting of a Global Structure Layout Tree and a Component Parameterization schema. 

      \item We develop a render-guided visual feedback and symbolic repair mechanism for recovering global facade structure from incomplete and non-rectified observations. 

      \item We develop a component-level inverse parameterization and deterministic compilation pipeline that assembles the recovered structure and local components into executable CGA programs for procedural 3D building asset generation.
    %


\end{enumerate}
\section{Related Work}

\paragraph{Procedural and inverse procedural modeling for architecture.}
In architectural reconstruction, procedural models instantiate executable structured programs that encode geometry through editable and executable symbolic rules. Classical grammar-based systems, such as Instant Architecture and CGA shape grammar, demonstrate the effectiveness of rule-based representations for buildings and urban scenes~\citep{wonka2003instant, muller2006procedural}. Inverse procedural modeling studies the reverse problem of recovering procedural structures and parameters from observations. The single-image building modeling system of \citet{nishida2018procedural} is closely related to our overall goal, as it recovers building mass, facade grammar, and window/door grammars from an image with a user-provided silhouette. Facade-level methods recover procedural facade definitions from rectified or segmented structures~\citep{muller2007image, koutsourakis2009single, teboul2011shape, wu2014inverse}, with FaçAID~\citep{facaid} learning a transformer-based mapping from segmented facades to procedural definitions. In our framework, such facade layout analysis corresponds mainly to the global structure recovery stage; we further ground repeated local tiles into procedural parameterization and assemble both levels into executable CGA programs from in-the-wild RGB images.

\paragraph{Visual program synthesis and neuro-symbolic reconstruction.}
More broadly, our task is related to visual program synthesis, where executable symbolic programs are inferred from visual observations. Early inverse-graphics systems such as SPIRAL~\citep{spiral} show that programs executed by a graphics engine can serve as structured image representations. Recent neuro-symbolic and program-based methods further extend this idea to 3D shape structures, CAD command sequences, and architectural abstractions, using domain-specific languages or program tokens as compact output spaces~\citep{shapeassembly, deepcad, arcpro}. Language-model-based inverse graphics methods such as IG-LLM~\citep{igllm} also suggest that structured scene code can be predicted from visual embeddings. These works show the benefit of moving visual reconstruction into executable program spaces, but direct full-program generation remains brittle when the output is non-unique, low-resource, and constrained by executability. Our method instantiates this direction for procedural facade reconstruction by predicting structured hierarchical neuro-symbolic representations that can be deterministically compiled into CGA programs.

\paragraph{Iterative visual reasoning and symbolic repair.}
Recent multimodal systems increasingly use intermediate visual evidence, tool invocation, or render-and-inspect loops to improve structured prediction. ICoT~\citep{icot} shows that inserting visual intermediates can strengthen fine-grained visual grounding beyond text-only rationales, while agentic systems such as GeoVista~\citep{geovista} integrate visual tools into multi-turn reasoning loops. Most closely related to our work, VIGA~\citep{viga} performs inverse graphics through a code-render-inspect loop, where rendered outputs provide visual evidence for subsequent symbolic edits. Our reflection stage adopts this principle for facade layout repair: the current Global Structure Layout Tree is rendered into a schematic layout, compared with RGB--mask evidence, and revised as CGA-compatible YAML under architectural priors.
\section{Method}

\textbf{Overview and problem formulation.} Given an in-the-wild building image, our goal is to recover an executable procedural facade representation for building asset generation. Since a single image provides only partial and view-dependent evidence of a building, we focus on recovering procedural facade descriptions for the dominant visible facade(s), which can then be attached to the corresponding faces of a simple mass model to instantiate an editable procedural building asset.

This formulation separates facade recovery from mass instantiation. The recoverable content from the image is the procedural structure and component configuration of visible facade regions, while the building asset is obtained by assigning the recovered facade programs to the corresponding sides of a default rectangular mass model. For unobserved sides, we use procedural completion rules or reuse recovered facade rules for visualization and downstream editing.

Following a coarse-to-fine architectural modeling strategy, we adopt a hierarchical two-level formulation. The first level recovers a facade-scale symbolic layout that captures the dominant organization of the facade and identifies representative tile regions. The second level takes these representative tiles as input and estimates tile-specific procedural parameters for component synthesis. The final executable program is obtained by deterministic assembly: the recovered structure layout is compiled into a global procedural scaffold, and the predicted tile-level codes are translated into executable component rules and inserted into the corresponding layout positions.

This decomposition is motivated by the ambiguity of direct image-to-code generation. Instead of asking a model to predict low-level procedural code in a single step, we cast the task as two more verifiable prediction problems: image-to-structure prediction and image-to-component-attributes prediction. By constraining neural prediction to structured intermediate representations and reserving executability for deterministic compilation, the framework provides a more stable bridge between image observations and procedural building modeling.
\begin{figure}
    \centering
    \includegraphics[width=\linewidth]{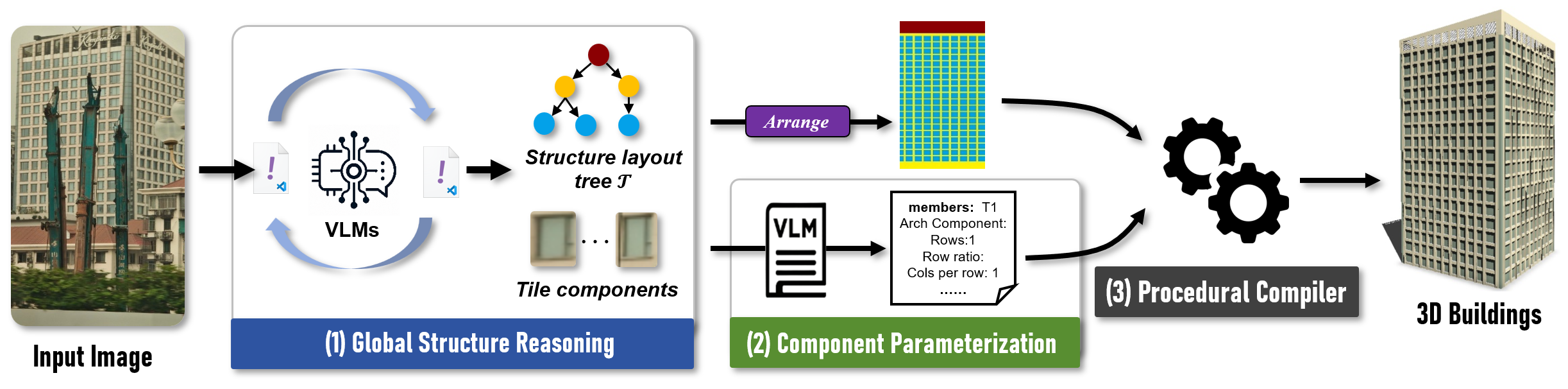}
    \caption{\textbf{Overview of our hierarchical neuro-symbolic framework.} The method decomposes facade reconstruction into a first level for symbolic layout recovery and a second level for tile procedural synthesis. The final CGA program is obtained by deterministic assembly of the global scaffold and the compiled tile components.}
    \label{fig:pipeline}
\end{figure}




\begin{figure}
    \centering
    \includegraphics[width=\linewidth]{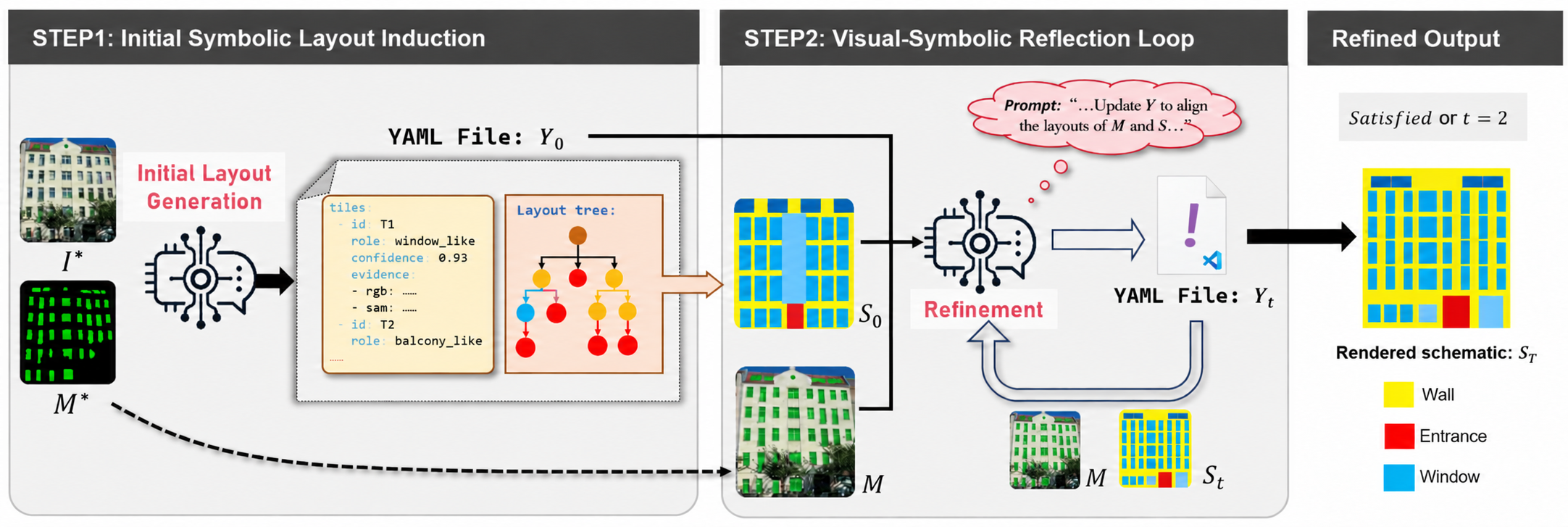}
    \caption{\textbf{Overview of global structure layout recovery via visual-symbolic reflection.}
    A fine-tuned Qwen-VL-7B model first induces an initial symbolic draft $Y_0$, including a tile set and a structure layout tree. The draft is then iteratively refined through a reflection loop driven by rendered schematics and RGB--mask overlay evidence, producing the final global structure layout output.}
    \label{fig:macro_reflection}
\end{figure}

\textbf{Global building structure reasoning.} The goal of this stage is to recover a \textit{\textbf{Global Structure Layout Tree $\mathcal{T}$}} from the input image. $\mathcal{T}$ is a rooted 2D hierarchy over the dominant facade plane. Here, the layout refers to a facade structural arrangement of floors and openings and other repeated tiles. It uses split and repeat operators to organize the layout, with parameters such as span, spacing, and axis direction specifying the normalized positions and extents of tile regions. This representation provides the global scaffold for downstream CGA generation. Detailed schema definitions are provided in Appendix~\ref{app:global_structure_layout}.

Given an input image $I$, we first localize the dominant facade support, extract the most relevant facade crop $I^*$, and use SAM3~\citep{sam3} to generate its corresponding mask $M^*$ for downstream recovery. The global structure recovery module is instantiated with
a fine-tuned Qwen-VL-7B model used in two instruction modes: layout induction and visual-symbolic reflection. The model is trained through a two-stage supervised fine-tuning procedure: initial layout induction followed by joint induction and refinement training.
Detailed training data construction and refinement trajectory generation are provided in Appendix~\ref{app:training_details}.

In the induction mode, the model constructs an initial symbolic draft of the $\mathcal{T}$ using a bottom-up, tile-first strategy,
\[
Y_0 = G_\theta(I^*, M^*).
\]
The resulting draft usually captures a plausible facade topology, but its spatial parameters and local boundaries may still be coarsely aligned with the image evidence.

To refine the draft, we introduce a render-guided visual feedback and symbolic repair mechanism implemented through an iterative reflection loop. Since the same facade may admit multiple valid YAML descriptions, direct comparison in the symbolic space is unreliable. We therefore render the current symbolic draft into a schematic layout and use the facade image together with segmentation-derived evidence as a weak visual reference for iterative symbolic revision. Specifically, we construct an overlay evidence map
\[
M = \Omega(I^*, M^*),
\]
which combines the facade crop with its corresponding SAM3 mask. At iteration $t$, the current symbolic draft $Y_t$ is rendered into a schematic layout image
\[
S_t = R(Y_t),
\]
where $R(\cdot)$ is a deterministic renderer and $S_t$ denotes the rendered visualization of the current layout draft. Here, $t=0$ corresponds to the initial layout draft $Y_0$ before reflection. In the reflection mode, the same fine-tuned Qwen-VL-7B model receives $\{M, Y_t, S_t\}$ and rewrites the current YAML draft into a revised symbolic program. This render-and-compare design makes the mismatch between the current layout and the visual evidence easier to identify and correct.

The reflection process is guided by soft architectural priors on structural regularity, semantic plausibility, and grammar validity, so that the recovered layout remains visually plausible and compilable. The loop terminates when sufficient agreement is reached or when a small fixed iteration budget is exhausted; in our implementation, we use $T_{\max}=2$. The final output of this stage consists of a refined $\mathcal{T}$ and a representative tile set derived from its leaf-level structure. The tree serves as the facade-scale scaffold for downstream procedural synthesis, while the tile set provides the local supports for tile-level procedural parameterization.

\textbf{Component-Level Inverse Parameterization.} After obtaining the global hierarchical tree of the building, we recover procedural descriptions for architectural components at the tile level. Each unique component is represented as a unified YAML-based set of procedural attributes. Instead of predicting raw CGA grammar, the model infers a constrained component-level parameterization defined by explicit architectural attributes (e.g., arch presence, row/column counts, and structural elements), which are sufficient for component synthesis. Detailed schema definitions are provided in Appendix~\ref{app:component_parameterization}. This representation serves as a spatial anchor for prediction, reducing ambiguity in component position and scale, while remaining CGA-compatible and casting under-specified local geometry as a structured parameterization problem. 

Given $\mathcal{T}$, we treat its leaf nodes as structural placeholders. The leaf-level layout provides a facade-scale estimate of component positions, while SAM3-derived local masks provide pixel-level evidence about component extents. By combining these two cues, we obtain representative tile supports for component parameterization.

Each grounded tile crop is encoded by a dual-stream visual backbone: a CLIP stream provides high-level semantic and stylistic cues, while a DINOv3 stream captures localized structural and geometric details. The fused visual features are projected into a shared decoder space and combined with language tokens for autoregressive generation of the component parameterization. A detailed architecture visualization is provided in Appendix~\ref{app:micro_architecture}.

\textbf{Deterministic procedural assembly and instantiation.} We adopt a CGA-style procedural construction paradigm following classical shape-grammar-based building modeling~\citep{wonka2003instant,muller2006procedural,muller2007image}, where production rules progressively refine building scopes from coarse facade regions to repeated tiles and local architectural details. 
In our implementation, the custom-designed compiler translates the global  structure layout tree into a CGA-compatible facade scaffold, and converts each component parameterization into a reusable component rule. 

After compilation, the recovered facade program can be attached to a building mass model to instantiate an editable 3D building asset in a procedural modeling engine. When multiple facades are visible in the input image, each recovered facade program can be assigned to the corresponding side of the mass model. For unobserved sides, we use simple procedural completion rules or reuse the recovered facade grammar for visualization and downstream editing. A simplified CGA-style pseudo-code example of this deterministic compilation and instantiation process is provided in Appendix~\ref{app:cga_pseudocode}.


\section{Experiments}

\begin{figure}
    \centering
    \includegraphics[width=\linewidth]{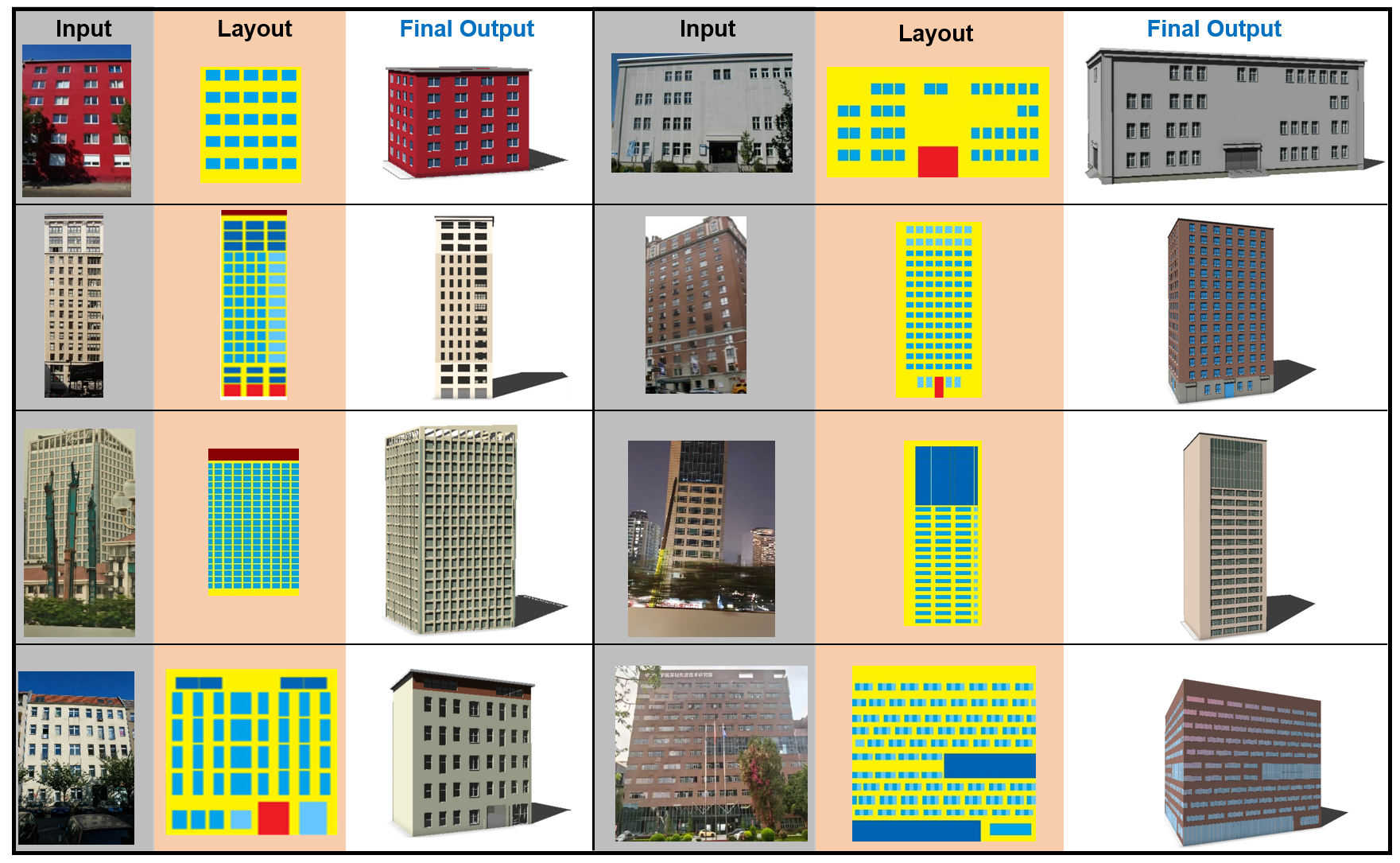}
    \caption{\textbf{End-to-end procedural reconstruction examples.}
    Each row shows two examples, each showing the input facade image, recovered layout, and final instantiated building asset. \textbf{For their detailed outputs, please refer to Fig.~\ref{fig:app_end2end_results_a} and~\ref{fig:app_end2end_results_b} in Appendix~\ref{app:end2end_results}.}}
    \label{fig:end2end_results}
\end{figure}

\subsection{Experimental Results} 

\textbf{End-to-end Result Showcase.} 
We first present representative end-to-end reconstruction results of the full framework. Given an in-the-wild building image, our method recovers a global structure layout scaffold and synthesizes tile-level procedural components, which are then deterministically assembled into an executable CGA program. The resulting representation can be rendered as a reconstructed procedural facade and further instantiated as a simple editable 3D building asset.

Fig.~\ref{fig:end2end_results} shows representative examples, including the input image, facade structure layout, and final procedural reconstruction. These results illustrate the overall goal of our framework: recovering not only a visually aligned facade layout, but also an executable procedural representation with reusable local architectural components. Additional end-to-end results and local details are provided in Appendix~\ref{app:end2end_results}.

\textbf{Effectiveness of Global Structure Reasoning.}
We evaluate whether the proposed global structure reasoning can recover a meaningful facade-scale structural scaffold from RGB images. The evaluation covers both regular facades, where layout structures are relatively clear, and challenging in-the-wild facades, where occlusion, perspective distortion, and incomplete evidence make structural recovery more difficult. 

\begin{figure}[t]
    \centering
    \includegraphics[width=\linewidth]{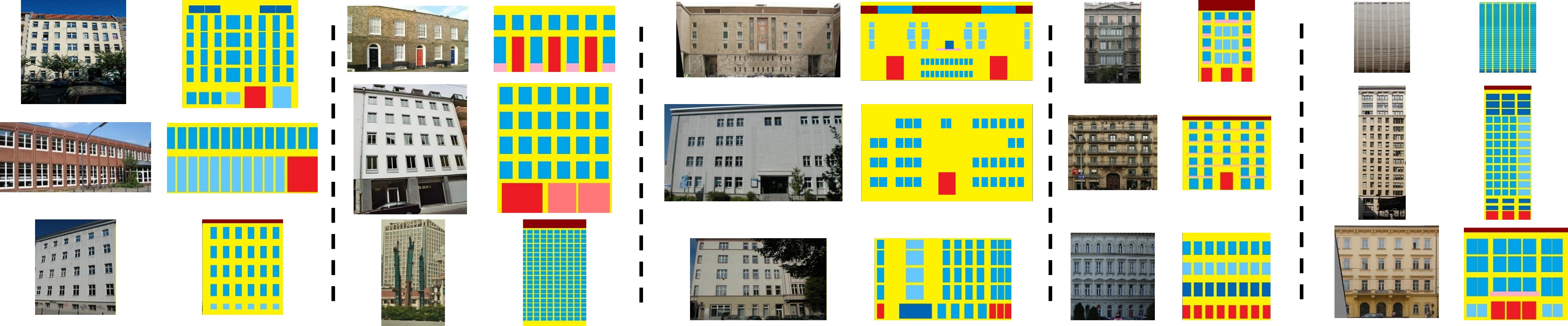}
    \caption{\textbf{Challenging facade reconstruction examples.}
    Representative examples from eTRIMS and in-the-wild images with corresponding recovered global structure layouts.}
    \label{fig:app_challenging_results}
\end{figure}

We further evaluate global structure recovery on challenging examples from eTRIMS and real-world imagery. These images include perspective distortion, partial occlusion, nonuniform spacing, and incomplete visual evidence. Since accurate procedural ground-truth layouts are generally unavailable for such cases, we focus on whether the recovered layouts preserve the dominant facade organization and remain procedurally meaningful. Additional challenging cases are provided in Fig.~\ref{fig:app_challenging_results}. 
Extended reflection-iteration results are reported in Table~\ref{tab:app_reflection_iterations} of Appendix~\ref{app:reflection_examples}.

\begin{figure}
    \centering
    \includegraphics[height=0.35\textheight,keepaspectratio]{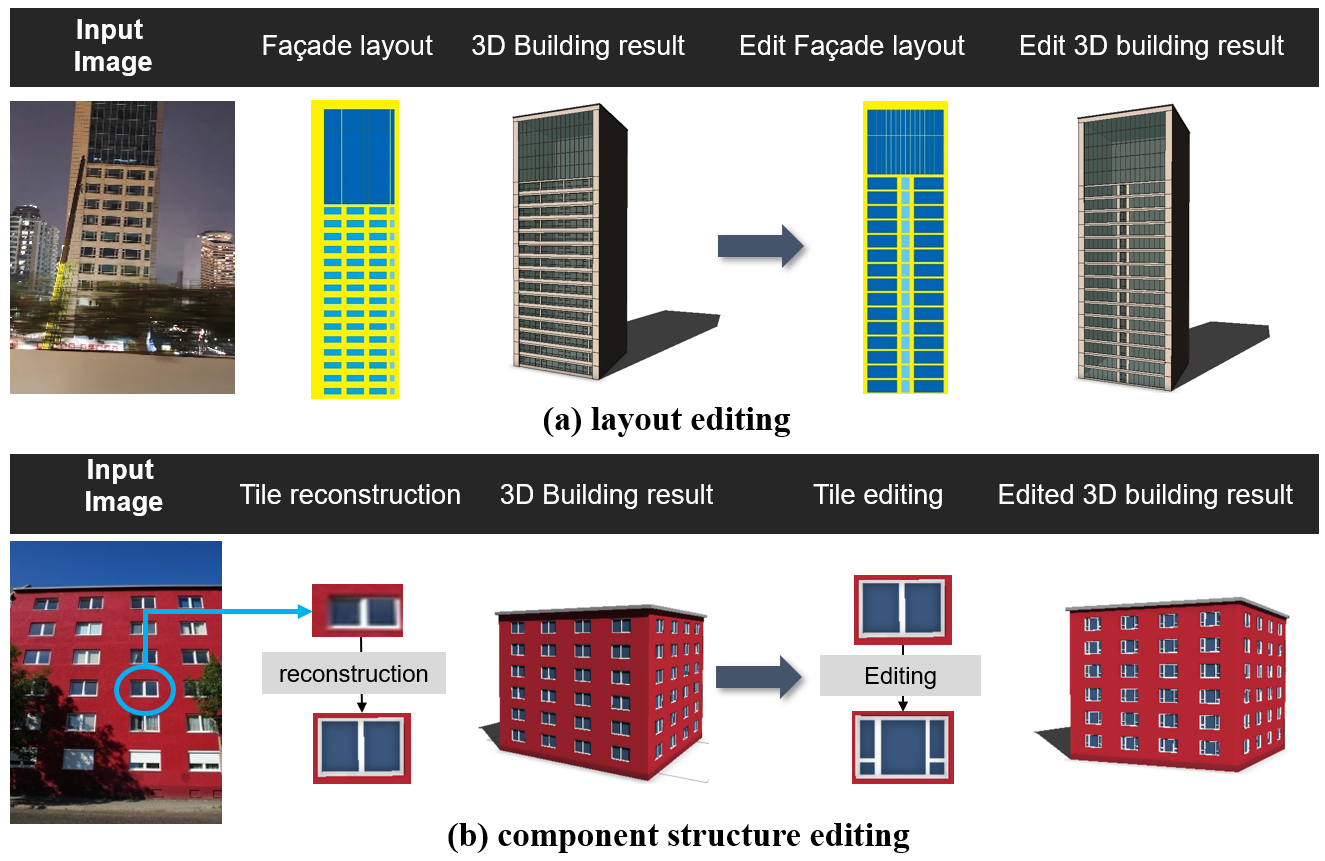}
    \caption{\textbf{Controllability and editability of the recovered procedural building representation.}
    Top: editing the recovered facade layout leads to a corresponding change in the instantiated 3D building result, demonstrating global structural controllability. 
    Bottom: editing a representative reconstructed tile produces updated repeated components in the final 3D building result, demonstrating local component-level editability. 
    Together, these examples show that the recovered procedural representation supports both facade-scale structural editing and local repeated-component editing.}
    \label{fig:tile_editability}
\end{figure}

\textbf{Effectiveness of Component Parameterization.}
We evaluate component parameterization using attribute-level accuracy on real-world tile crops from CMP~\cite{tylecek2013cmpfacade} and eTRIMS~\cite{korc2009etrims}, as well as procedurally rendered synthetic tiles. Real-world evaluation uses 107 held-out crops, with training and test partitions separated by building identity. Data construction is described in Appendix~\ref{app:training_details}, with the real-tile split and leakage checks detailed in Appendix~\ref{app:real_tile_split}.

For geometric attributes, we report exact-match accuracy over discrete presets of normalized values. Attribute discretization and metric definitions, together with continuous-error evaluation on synthetic test tiles, are detailed in Appendix~\ref{app:attribute_metrics}. A separate comparison of synthetic-only and synthetic-plus-real training on the same held-out real test set is reported in Appendix~\ref{app:synthetic_to_real_transfer}.

\begin{table}
\centering
\caption{\textbf{Accuracy of representative component (tile) attributes on real-world and synthetic datasets.}
W and E denote window-like and entrance-like components. Full per-attribute results are provided in Appendix~\ref{app:tile_level_quant_results}.}
\label{tab:tile_attr_acc}
\small
\setlength{\tabcolsep}{4pt}
\begin{tabular}{lccccc}
\toprule
Data & W-Arch$\uparrow$ & W-Sill$\uparrow$ & W-Row/Col$\uparrow$ & E-Stair$\uparrow$ & E-Handle$\uparrow$ \\
\midrule
Real-world & 0.98 & 0.92 & 0.89 & 0.95 & 0.96 \\
Synthetic  & 0.99 & 0.96 & 0.92 & 0.98 & 0.97 \\
\bottomrule
\end{tabular}
\end{table}

Table~\ref{tab:tile_attr_acc} shows that the model achieves strong attribute-level accuracy on both synthetic and real-world tile sets, with a moderate gap between the two settings. 
W and E denote window-like and entrance-like tiles, respectively. 
These results indicate that the component parameterization predictor can recover key local procedural attributes from real image evidence while benefiting from the controllability and coverage of synthetic supervision. Detailed per-attribute results are provided in Appendix~\ref{app:tile_level_quant_results}.

\paragraph{Bonus Feature: Editability.}

One major advantage of our method lies in its ability to produce 3D results with separable structures, enabling seamless integration into game engines and facilitating user-driven post-editing.
Fig.~\ref{fig:tile_editability} shows two examples of user edits on our reconstructions at both the facade and tile levels.
The modification on the global facade layout enables changes to the overall structure of the instantiated 3D building (see Fig.~\ref{fig:tile_editability}(a)). In contrast, editing the parameters of local tiles enables updates to all corresponding repeated components in the final building (see Fig.~\ref{fig:tile_editability}(b)). This demonstrates that our output is not merely a static mesh or image, but an executable and editable procedural representation that supports both global structural control and local component-level manipulation.

\subsection{Comparison}

We conduct comparison studies at two levels: global structure recovery and Component Parameterization. 

For \textbf{global facade-structure recovery}, 
we compare our recovered global facade structures with representative examples reported in FaçAID. Since FaçAID assumes segmented facade input, while our method starts from RGB facade images, and since no public implementation or unified executable evaluation protocol is available, we treat this comparison as a qualitative layout-level reference rather than a direct benchmark under identical inputs. Fig.~\ref{fig:facaid_comparison} shows a representative comparison, and additional visual comparisons are provided in Appendix~\ref{app:additional_macro_results}.

\begin{figure}
    \centering
    \includegraphics[width=0.8\linewidth]{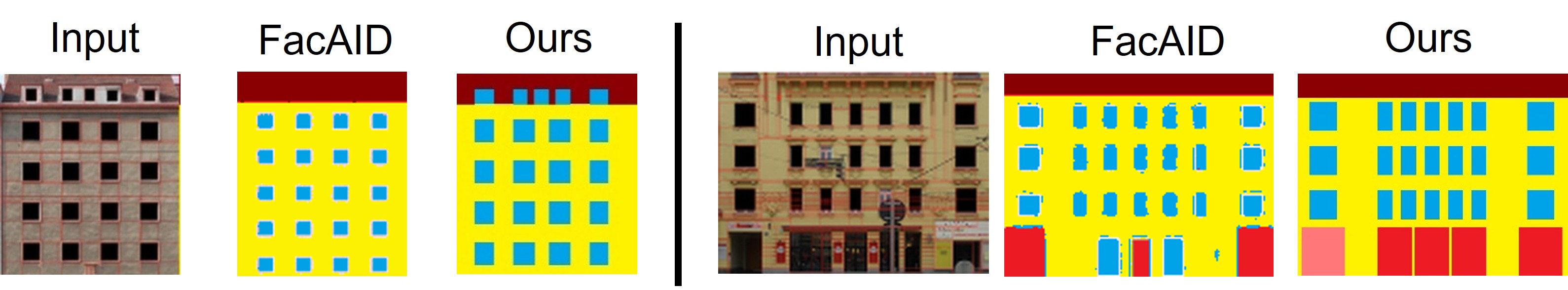}
    \caption{\textbf{Representative comparison with FaçAID.}
    From left to right are the input facade image, the layout result reported by FaçAID, and our recovered global facade structure.}
    \label{fig:facaid_comparison}
\end{figure}

To further complement this qualitative comparison, we conduct a small user preference study. Participants are shown the input facade image and two anonymized layout results, one from FaçAID and one from our method, and are asked which result better matches the facade structure. Table~\ref{tab:user_study_facaid} reports the percentage of preferences for FaçAID, ours, and tie, where tie indicates that the two results are judged comparable or hard to distinguish. The study involved 15 participants and 180 randomized blinded comparisons. Participant information, the evaluation protocol, full instructions, vote counts, and exploratory statistical results are provided in
Appendix~\ref{app:user_study}.

\begin{table}[t]
\centering
\caption{\textbf{User preference study comparing FaçAID and our method in terms of facade reconstruction fidelity.} Participants selected their preferred result, and we report the percentage of votes. '\textit{Tie}' indicates the proportion of cases where both results were judged to be of comparable quality.}
\label{tab:user_study_facaid}
\small
\begin{tabular}{l|ccc}
\toprule
Methods: & FaçAID & Ours & Tie \\
\midrule
Vote Proportion: & 30\% & \textbf{59}\% & 11\% \\
\bottomrule
\end{tabular}
\end{table}

Next, we evaluate the role of the \textbf{Component Parameterization} by comparing three component-level generation strategies: direct low-level code generation, direct component parameterization prediction, and our fine-tuned predictor. Execution success is reported for all methods, while attribute-level accuracy is reported only for methods that explicitly predict structured tile attributes. Representative qualitative examples are provided in Appendix~\ref{app:direct_generation_examples}.

\begin{table}[!htbp]
\centering
\caption{\textbf{Comparison of tile-level procedural generation strategies.}
Attribute accuracy is reported only for methods that explicitly predict tile-level structured attributes.}
\label{tab:tile_level_comparison}
\small
\setlength{\tabcolsep}{4pt}
\resizebox{\linewidth}{!}{
\begin{tabular}{lcccccc}
\toprule
Method & Exec. Succ.↑ & Arch Acc.↑ & Row/Col Acc.↑ & Sill Acc.↑ & Stair Acc.↑ & Handle Acc.↑\\
\midrule
Direct Code (Gemini 3.1 Pro Preview)          & 0.68 & N/A & N/A & N/A & N/A & N/A\\
Direct Middle Code (Gemini 3.1 Pro Preview)   & 0.95 & 0.94 & 0.88 & 0.88 & 0.96 & 0.91\\
Ours                                          & \textbf{0.98} & \textbf{0.99} & \textbf{0.90} & \textbf{0.93} & \textbf{0.96}& \textbf{0.96} \\

\bottomrule
\end{tabular}
}
\end{table}

Table~\ref{tab:tile_level_comparison} reports execution success and representative attribute accuracy on objective tile-level properties. Compared with the stronger Gemini 3.1 Pro Preview baseline, direct middle-code generation substantially improves executability over direct code generation, indicating that structured intermediate representations reduce the ambiguity of procedural program generation. Our task-specific predictor further improves execution reliability and fine-grained architectural attribute recovery.

Lastly, we compare our \textbf{full framework} with a direct generation baseline, where a general LLM/VLM is prompted to generate the entire CGA program from the input facade image in a single pass. We evaluate this comparison from both the layout and executable-code levels using Tree Repair Cost (TRC) and execution success rate. For the direct generation baseline, the layout tree used for TRC evaluation is recovered from the generated CGA program. TRC is computed on the layout structure and measures the expert-validated structural repair cost under a fixed editing protocol. Specifically, annotators edit the predicted layout tree through predefined atomic operations until its rendered schematic expresses the dominant facade organization. Each insertion, deletion, or modification of a structural node counts as one edit. Therefore, TRC reflects the amount of manual structural correction required by a prediction, rather than a global optimum over all possible tree representations. The detailed annotation interface and repair protocol are provided in Appendix~\ref{app:trc_protocol}.

Table~\ref{tab:direct_generation} reports execution success and symbolic repair cost, and representative qualitative examples are provided in Appendix~\ref{app:direct_generation_examples}.

\begin{table}[t]
\centering
\caption{Comparison with direct end-to-end LLM/VLM generation.}
\label{tab:direct_generation}
\small
\setlength{\tabcolsep}{8pt}
\begin{tabular}{lcc}
\toprule
Method & Exec. Succ.$\uparrow$ & TRC$\downarrow$ \\
\midrule
Direct LLM (Gemini 3.1 Pro Preview) & 0.94 & 1.83 \\
Ours                        & \textbf{0.98} & \textbf{1.47} \\
\bottomrule
\end{tabular}
\end{table}

\subsection{Ablation}

We first evaluate the \textbf{effect of visual-symbolic reflection} on global facade-structure recovery. Starting from the initial symbolic draft, the model iteratively revises the layout using rendered feedback and segmentation-derived visual evidence. In addition to TRC and execution success rate, we report count-based structural accuracy: floor accuracy is computed at the facade level, while window and door count accuracy are computed at the floor level by comparing the predicted and target element counts.

\begin{table}[!htbp]
\centering
\caption{Ablation of visual-symbolic reflection on structure layout recovery.}
\label{tab:reflection_effect}
\small
\setlength{\tabcolsep}{5pt}
\begin{tabular}{lccccc}
\toprule
Model & Win Cnt. Acc.$\uparrow$ & Floor Cnt. Acc.$\uparrow$ & Door Cnt. Acc.$\uparrow$ & TRC$\downarrow$ & Exec. Succ.$\uparrow$ \\
\midrule
Initial Layout & 0.61 & 0.83 & 0.69 & 4.32 & 0.98 \\
Iteration 1    & 0.77 & 0.85 & 0.78 & 2.01 & 0.98 \\
Iteration 2    & \textbf{0.80} & \textbf{0.86} & \textbf{0.81} & \textbf{1.47} & \textbf{0.99} \\
\bottomrule
\end{tabular}
\end{table}

Table~\ref{tab:reflection_effect} reports the main reflection results up to the default setting $T_{\max}=2$, showing that reflection improves count-level structural consistency and reduces TRC while maintaining high global structure recovery compilability. Most of the improvement is obtained within the first two iterations, after which the gains become marginal. We therefore use $T_{\max}=2$ in the main experiments. Additional qualitative examples and an extended analysis with more reflection iterations are provided in Appendix~\ref{app:reflection_examples}.

We further ablate the \textbf{visual stream} by comparing CLIP-only, DINOv3-only, and CLIP+DINOv3 encoders using grouped semantic-presence and structural attributes for window-like and entrance-like tiles. The combined encoder achieves the best overall structural accuracy, with full grouped results provided in Appendix~\ref{app:tile_ablation_results}.

\FloatBarrier
\subsection{Limitations and Discussion}

Our framework targets approximately planar visible facades, while unseen geometry is completed for asset instantiation rather than reconstructed from image evidence. The current component schema supports simple parametric window arches, but not entrance arches or complex decorative assemblies. Severe perspective distortion and occlusion may still affect layout recovery.

Our experiments cover the considered facade sources and include building-disjoint real-tile evaluation, but controlled city-disjoint and style-disjoint evaluation remains future work. More expressive massing and component representations, together with additional geometric evidence, could extend the framework to more complex architecture.
\section{Conclusion}

We presented a hierarchical neuro-symbolic framework for visual-to-executable procedural reconstruction of buildings from in-the-wild images. The key idea is to bridge visual facade understanding and procedural generation through structured intermediate representations: a global structure layout tree captures facade-scale organization, while component parameterizations describe local architectural elements. Together with render-guided visual-symbolic reflection and deterministic CGA compilation, this design avoids direct low-level code generation and produces editable, executable procedural facade programs.

Experiments show that the proposed representation-aligned formulation recovers meaningful facade structures on both regular and challenging in-the-wild images, improves procedural executability over direct LLM/VLM generation baselines, and preserves fine-grained component details through tile-level parameter prediction. These results suggest that structured intermediate representations provide a practical path toward robust image-based procedural building reconstruction.

\begin{ack}
This work was supported in part by the Guangdong Major Project of
Basic and Applied Basic Research (Grant No. 2023B0303000016) and the
Shenzhen Science and Technology Program
(Grant No. CJGJZD20240729141906008).
\end{ack}

\bibliographystyle{plainnat}
\bibliography{reference}

\appendix

\section{More Reconstruction Results}
\label{app:end2end_results}

\begin{figure}[htbp]
    \centering
    \includegraphics[width=0.92\linewidth]{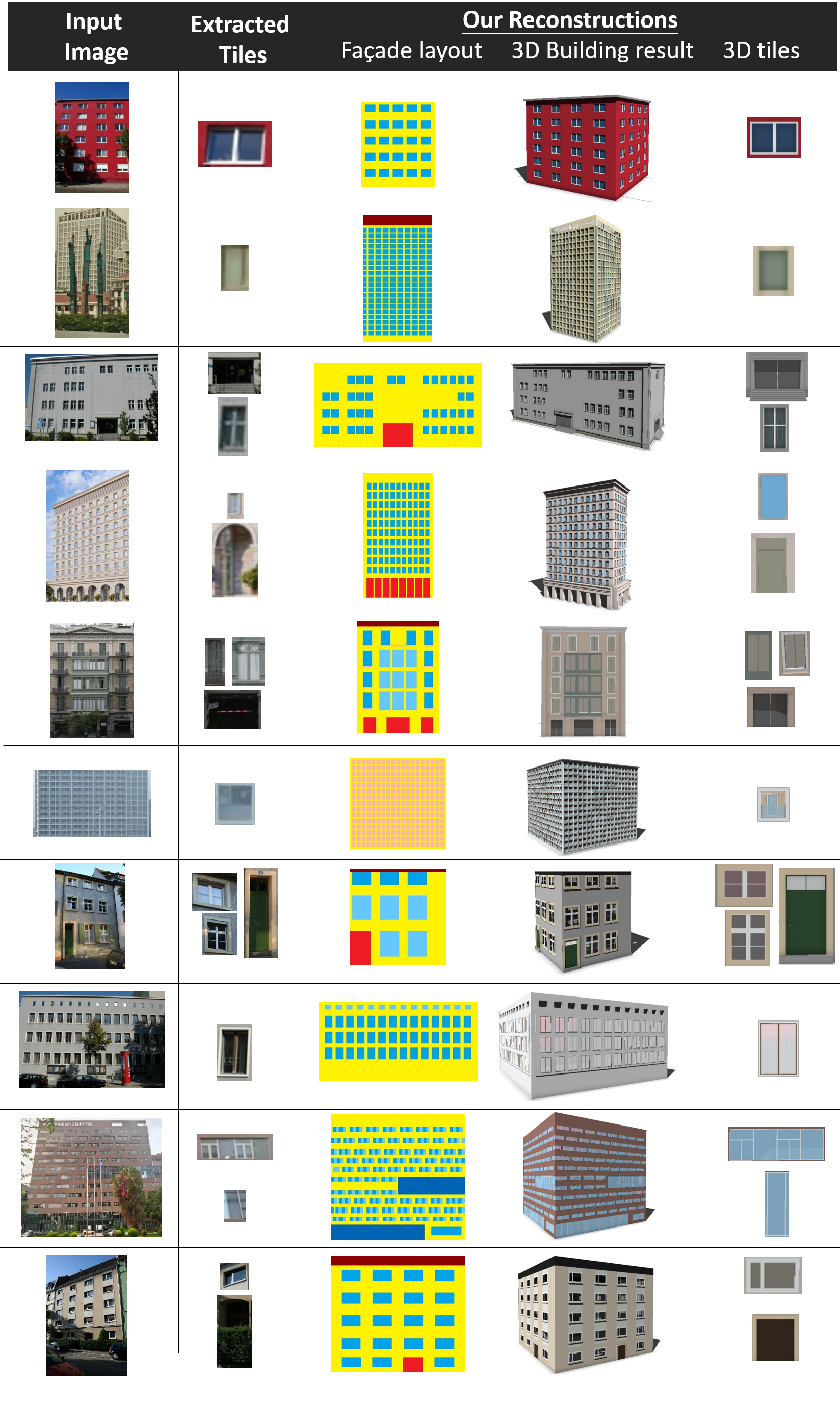}
    \caption{\textbf{Additional end-to-end procedural reconstruction results - 1.}
    Each example shows the input building image and the final procedural reconstruction generated by our full framework.}
    \label{fig:app_end2end_results_a}
\end{figure}

\begin{figure}[htbp]
    \centering
    \includegraphics[width=0.92\linewidth]{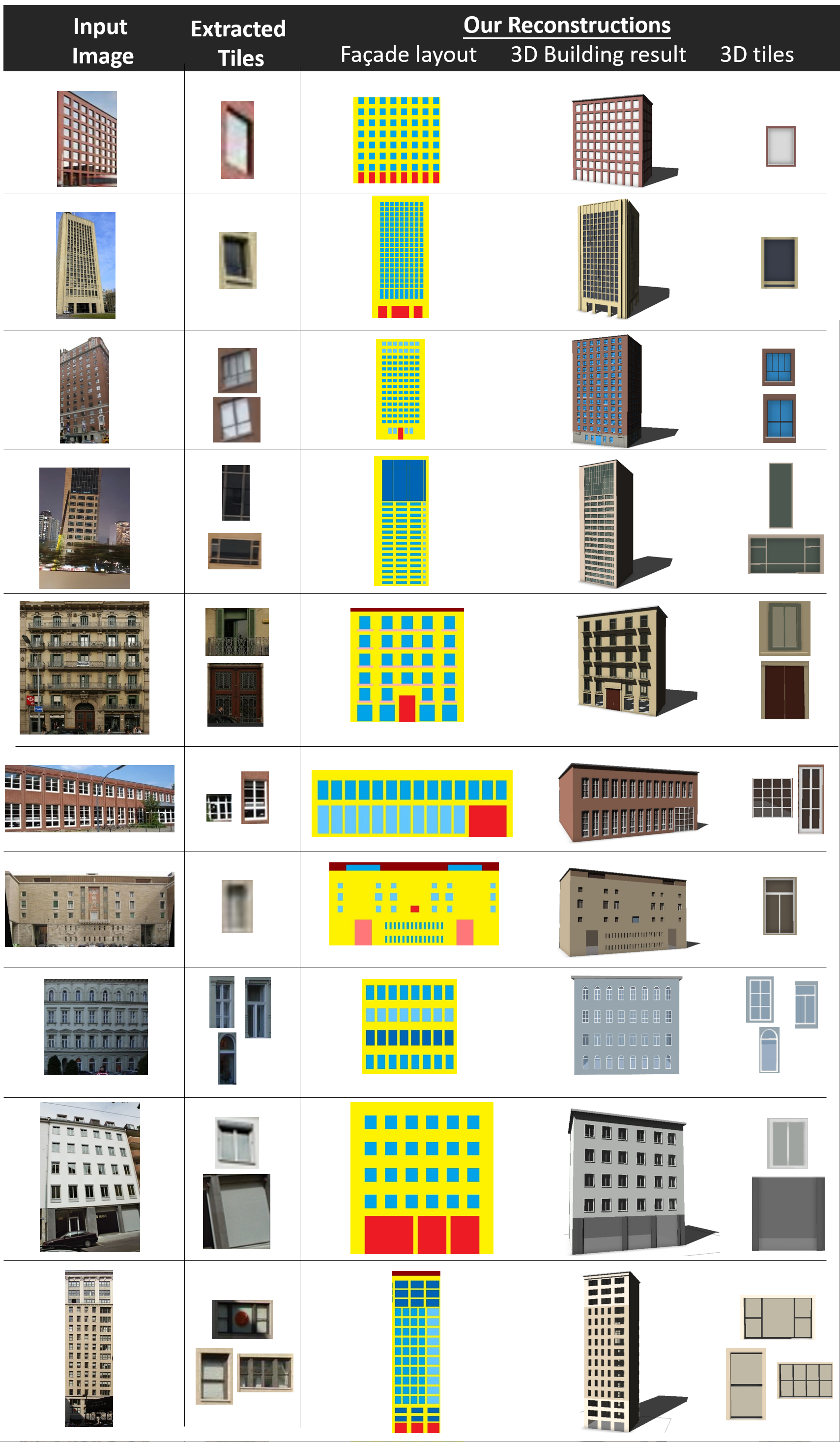}
    \caption{\textbf{Additional end-to-end procedural reconstruction results - 2.}
    Each example shows the input building image and the final procedural reconstruction generated by our full framework.}
    \label{fig:app_end2end_results_b}
\end{figure}

Fig.~\ref{fig:app_end2end_results_a} and \ref{fig:app_end2end_results_b} provide additional end-to-end reconstruction results. Each example shows the input building facade image together with the final procedural reconstruction produced by our framework. These results complement the main-paper examples and further demonstrate the ability of the proposed method to combine global structure layout recovery and component-level procedural parameterization into a unified executable facade representation.

\section{Detailed Representation Schemas}
\label{app:representation_schema}

\begin{figure}[t]
    \centering
    \includegraphics[width=\linewidth]{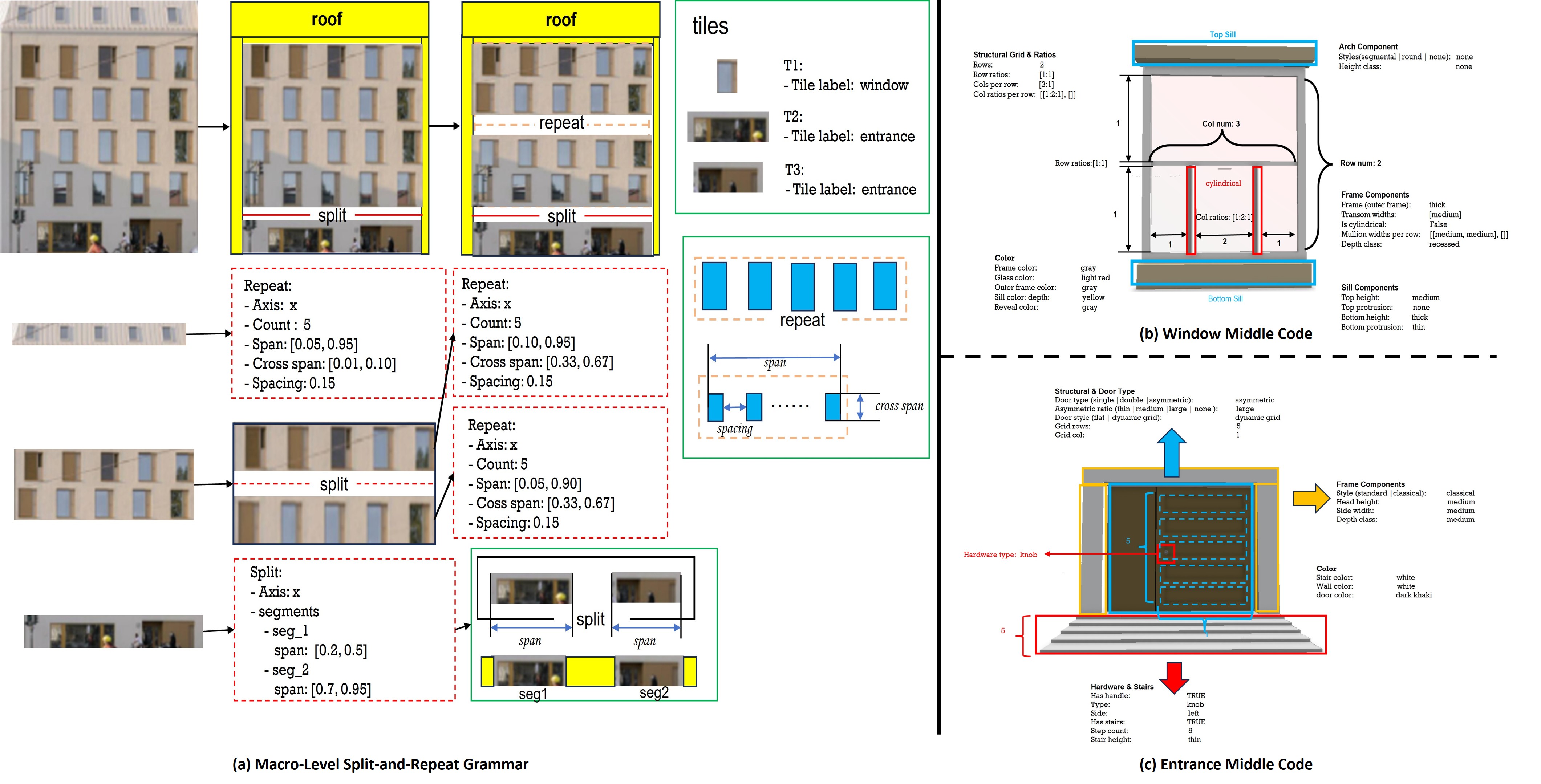}
    \caption{\textbf{Detailed dual-tier symbolic representation.}
    The left panel shows the grammar used to describe facade-scale organization. The right panels show representative component parameterization for window-like and entrance-like components.}
    \label{fig:appendix_dual_ir}
\end{figure}

This section provides additional details of the hierarchical two-level representation used in our framework. Fig.~\ref{fig:appendix_dual_ir} illustrates the global structure layout schema for split-and-repeat facade organization and the component-level parameterization schemas for representative window and entrance components.

\subsection{Global structure layout schema.}
\label{app:global_structure_layout}
The global structure level representation describes the dominant facade plane as a hierarchical split-and-repeat layout. Each region is represented by a symbolic node with a normalized spatial domain and a node type, such as split, repeat, or leaf. Split nodes partition a parent region into ordered child regions, while repeat nodes describe repeated tile instances along a selected axis. Leaf nodes serve as placeholders for representative tiles, which are later grounded and replaced by compiled component rules.

\subsection{Component parameterization schema.}
\label{app:component_parameterization}
The component representation abstracts each unique tile into a structured set of procedural attributes rather than raw CGA code. For window-like tiles, the schema includes local grid structure, frame configuration, arch and sill attributes, depth-related properties, and appearance parameters. For entrance-like tiles, the schema includes door structure, frame configuration, stairs, hardware, and appearance attributes. These attributes are designed to be directly translatable into CGA component rules by the deterministic compiler.

\subsection{CGA-style Procedural Compilation Example}
\label{app:cga_pseudocode}

Listing~\ref{lst:cga_example} gives a simplified CGA-style pseudo-code example of our deterministic compilation process. 
The facade is first subdivided along the vertical axis into floor-level regions, repeated floors are expanded into horizontal tile slots, and each leaf tile invokes a parameterized component generator such as \texttt{Window} or \texttt{Entrance}. 
The compiled component rules are then scaled to their corresponding leaf scopes and assembled on the facade scaffold. 
Finally, the assembled facade is attached to a simple building mass model to produce an editable and executable procedural 3D building.

\begin{lstlisting}[caption={A simplified CGA-style pseudo-code example illustrating how global layout and component parameters are compiled into procedural facade rules.},label={lst:cga_example}]
0: building -> comp(f) {
       front : facade
       | back : Wall
       | left : facade
       | right: facade
}

1: facade -> Subdiv(Y, ground_h, 1r, top_h) {
       ground_floor | Repeat(Y, floor_h) { typical_floor } | top
   }

2: typical_floor -> Repeat(X, tile_w) {
       window_slot
   }

3: ground_floor -> Subdiv(X, 1r, door_w, 1r) {
       wall | entrance_slot | wall
   }

4: window_slot -> Window(
       rows=2, cols=3, frame=true, arch=false, sill=true
   )

5: entrance_slot -> Entrance(
       door_type=double, stairs=true, handle=true
   )
\end{lstlisting}

\section{Additional Visual-Symbolic Reflection Examples}
\label{app:reflection_examples}

We further analyze the effect of increasing the number of visual-symbolic reflection iterations. This study examines whether more reflection steps consistently improve the recovered global structure layout. As shown in Table~\ref{tab:app_reflection_iterations}, most of the improvement is obtained within the first two iterations: count-level structural accuracy improves substantially and TRC decreases sharply compared with the initial layout. Further iterations bring only marginal changes and may introduce small fluctuations in local element counts. Therefore, we use a small fixed reflection budget in the main experiments.

\begin{figure}[t]
    \centering
    \includegraphics[width=\linewidth]{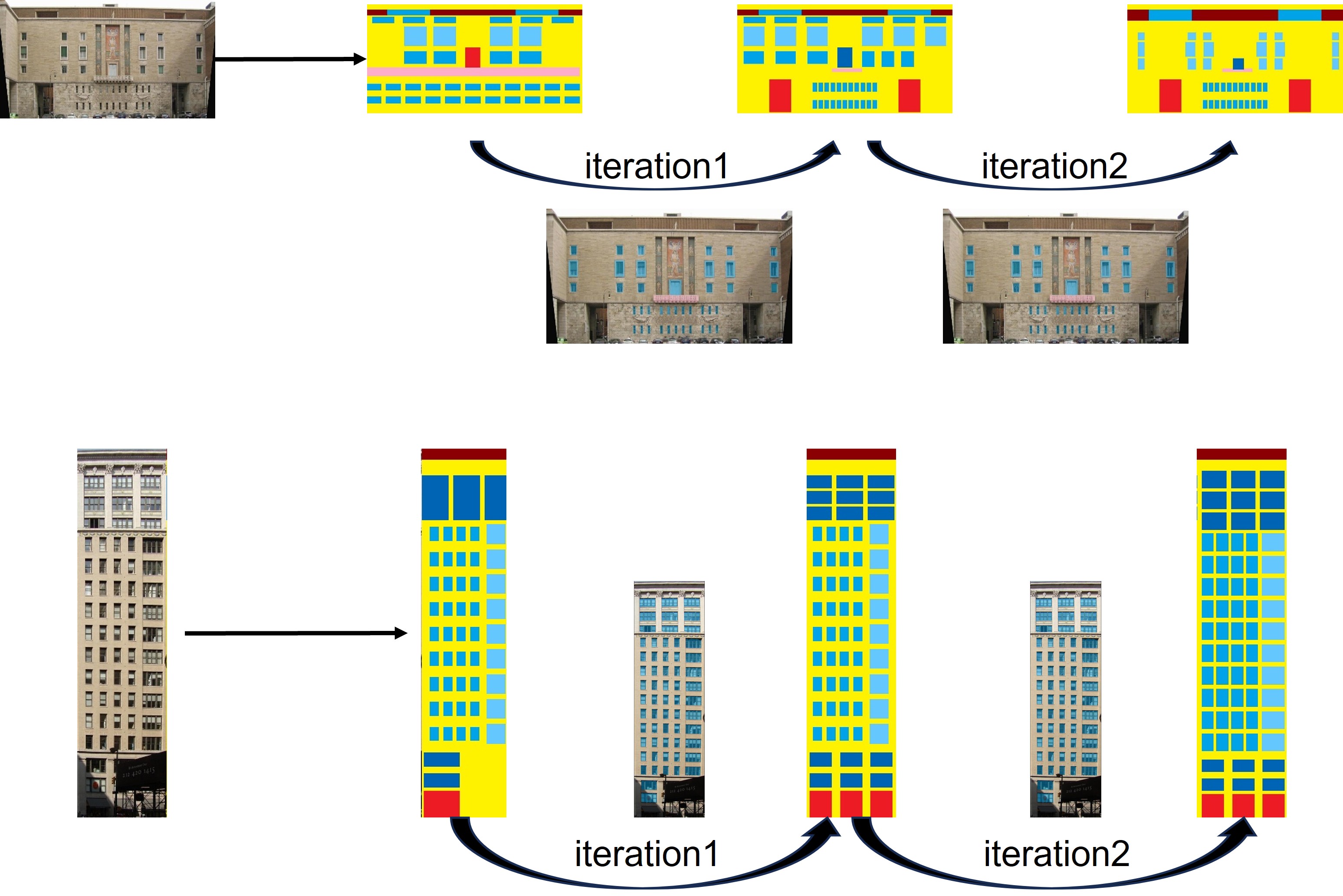}
    \caption{\textbf{Updated examples under the process of visual-symbolic reflection loop.}
    The examples show how the initial global structure layout draft is refined over reflection iterations using rendered schematic feedback and RGB--mask overlay evidence.}
    \label{fig:app_reflection_examples}
\end{figure}

Fig.~\ref{fig:app_reflection_examples} provides qualitative examples of the reflection process. The examples illustrate how the initial symbolic draft is iteratively refined through rendered layout feedback and RGB--mask overlay evidence.

\begin{table}[t]
\centering
\caption{Effect of increasing the number of visual-symbolic reflection iterations.}
\label{tab:app_reflection_iterations}
\small
\setlength{\tabcolsep}{7pt}
\begin{tabular}{lccccc}
\toprule
Model & Acc (Win)$\uparrow$ & Acc (Floor)$\uparrow$ & Acc (Door)$\uparrow$ & TRC$\downarrow$ & Exec. Succ.$\uparrow$ \\
\midrule
Initial Layout & 0.61 & 0.83 & 0.69 & 4.32 & 0.98 \\
Iteration 1    & 0.77 & 0.85 & 0.78 & 2.01 & 0.98 \\
Iteration 2    & \textbf{0.80} & 0.86 & \textbf{0.81} & 1.47 & \textbf{0.99} \\
Iteration 3    & \textbf{0.80} & \textbf{0.87} & 0.80 & \textbf{1.46} & \textbf{0.99} \\
Iteration 4    & 0.78 & \textbf{0.87} & 0.79 & 1.48 & \textbf{0.99} \\
\bottomrule
\end{tabular}
\end{table}

\section{Training and Implementation Details}
\label{app:training_details}

\paragraph{Qwen-VL-7B fine-tuning.}
We fine-tune a Qwen-VL-7B model for global structure layout recovery using both real and synthetic facade data. Each training example pairs a facade image and segmentation-derived evidence with a normalized orthogonal layout tree represented in YAML. For real data, we use in-the-wild facade crops without explicit perspective rectification and manually annotate their layouts. For synthetic data, we augment the training inputs by rendering procedurally generated facades from different viewpoints, pairing each rendered image with its corresponding layout derived from the procedural rules. These pairs provide supervision for mapping view-dependent image
evidence to canonical facade structures.

To account for the non-uniqueness of procedural layout representations, we annotate each real facade with at least two valid layout strategies when possible. These alternative YAML layouts may differ in grouping, split order, or repeat abstraction, while still describing the same dominant facade organization. This design avoids forcing the model to imitate a single arbitrary tree representation and better reflects the multi-solution nature of procedural facade modeling. The fine-tuned Qwen-VL-7B model is used in two instruction modes during inference: initial layout induction and visual-symbolic reflection.

\paragraph{Two-stage supervised fine-tuning.}
We first fine-tune the model on layout-induction examples to establish the image-to-layout mapping. Starting from this checkpoint, we jointly fine-tune it on induction and refinement examples distinguished by task-specific instructions. The resulting checkpoint is shared by the two instruction modes at inference time.

\paragraph{Visual-symbolic refinement training.}

To train the refinement capability, we further construct supervised repair
trajectories based on valid facade layouts. Given an accepted layout
representation $Y$, we generate intermediate corrupted states by applying
schema-valid perturbations at different levels.

Starting from an accepted layout $Y_{\mathrm{valid}}$, we first perturb normalized spans and spacing to obtain $Y_{\mathrm{fine}}$, and then perturb topology and discrete counts to obtain $Y_{\mathrm{coarse}}$. Thus, data corruption follows $Y_{\mathrm{valid}}\rightarrow Y_{\mathrm{fine}}\rightarrow Y_{\mathrm{coarse}}$, whereas supervised repair follows the reverse direction:

\begin{equation}
Y_{\mathrm{coarse}}
\rightarrow Y_{\mathrm{fine}}
\rightarrow Y_{\mathrm{valid}}.
\label{eq:app_refinement_trajectory}
\end{equation}

During refinement training, we supervise the VLM on successive steps of the coarse-to-fine repair trajectory. Each example uses the facade image, segmentation-derived evidence, the current YAML layout, and a schematic rendered from that layout as input, with the next repaired layout as the prediction target.

For synthetic data, refinement trajectories are automatically generated using
schema-valid and renderable perturbations. For real facade images, we manually
construct repair trajectories by first correcting structural/count errors and
then refining spatial parameters. When multiple valid decompositions exist,
alternative executable layouts are also included as valid examples rather than
forcing the model toward a single tree serialization.

\paragraph{Training configuration.}
We fine-tune the layout model using LoRA and AdamW, with a learning rate of $10^{-5}$, weight decay of $0.1$, and $\beta_2=0.95$. Training uses BF16 precision, a per-device batch size of 2, 8 gradient-accumulation steps, a maximum sequence length of 2048, and 5 epochs per training run. We use a LoRA rank of 64, a scaling parameter of 16, and dropout of 0.05. Gradient checkpointing and DeepSpeed ZeRO-2 are enabled.

\paragraph{Tile-level predictor training data.}
We construct synthetic and real tile datasets for component parameterization.
Each sample pairs a component crop with structured procedural attributes
under the schema in Appendix~\ref{app:component_parameterization}.
Data construction and the building-disjoint real-tile split are detailed
in Appendix~\ref{app:tile_datasets}.

\subsection{Component Parameterization Architecture}
\label{app:micro_architecture}

\begin{figure}[t]
    \centering
    \includegraphics[width=\linewidth]{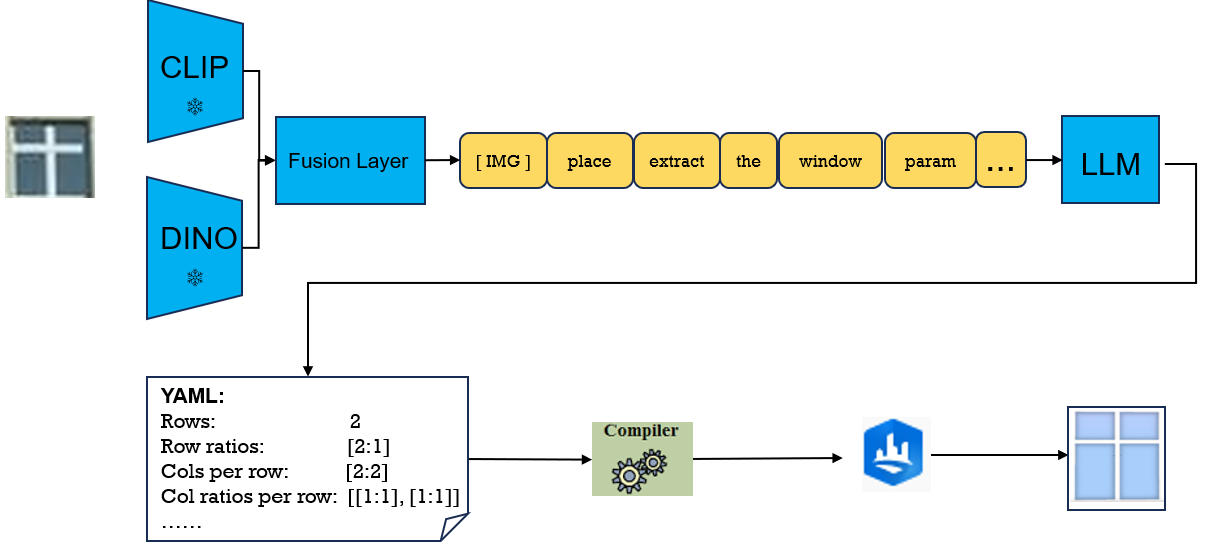}
    \caption{\textbf{Our component parameterization architecture.}
    Representative tile crops are encoded by a dual-stream visual backbone and decoded into tile-specific YAML middle code, which is then deterministically compiled into executable CGA snippets.}
    \label{fig:app_micro_tile}
\end{figure}

Fig.~\ref{fig:app_micro_tile} shows the detailed architecture of the component parameterization predictor. Representative tile crops are encoded by the dual-stream CLIP and DINOv3 backbone, fused into a shared decoder space, and decoded into component parameterization for deterministic procedural compilation.

\subsection{Tile Dataset Construction and Splits}
\label{app:tile_datasets}
\label{app:real_tile_split}

\paragraph{Synthetic tile construction.}
Synthetic tiles are rendered using parameterized procedural generators for common window-like and entrance-like components. The generators vary grid structure, frames, window arches, sills, stairs, hardware, and appearance attributes. Each rendered image is paired with the corresponding component-parameterization annotation derived from the generator parameters. The original continuous CGA parameters are also available for evaluating geometric attributes on synthetic tiles. Both synthetic and real annotations follow the schema in Appendix~\ref{app:component_parameterization}.

\paragraph{Real tile collection and annotation.}
Real window and entrance crops are collected from the CMP Facade
Database~\cite{tylecek2013cmpfacade} and eTRIMS~\cite{korc2009etrims}
and manually annotated with structured component attributes. Before extracting crops, we group all images of the same building, including multiple viewpoints, repeated captures, and overlapping images. We then segment window and entrance instances from the images in each building group and extract their component crops.

\paragraph{Redundancy removal and building-level partitioning.}
Within each building group, we extract DINOv2 features for the component crops and remove highly similar crops using a cosine-similarity threshold of $0.85$. This removes redundant instances, such as nearly identical windows from the same facade or duplicate observations of the same opening. DINOv2 is used here for dataset redundancy checks, whereas the component predictor uses CLIP and DINOv3.

The train/test partition is constructed at the building-group level. Building groups are randomly assigned until the retained crop counts are approximately balanced between the two partitions. All viewpoints, facades, and component crops from one building remain in the same partition. Highly similar cross-split pairs are additionally inspected to remove duplicated image content or same-building overlap. Table~\ref{tab:app_real_tile_split} summarizes the resulting 207 crops from 67 building groups: 100 training crops and 107 test crops.

\begin{table}[htbp]
\centering
\caption{Building-disjoint partition of the real tile dataset collected
from CMP and eTRIMS. The 207 crops comprise both training and test data.}
\label{tab:app_real_tile_split}
\small
\begin{tabular}{lrrr}
\toprule
Split & Tiles & Building groups & Avg. tiles/group \\
\midrule
Training & 100 & 31 & 3.23 \\
Test     & 107 & 36 & 2.97 \\
\midrule
Total    & 207 & 67 & 3.09 \\
\bottomrule
\end{tabular}
\end{table}

\section{Additional Tile-Level Quantitative Results}
\label{app:tile_level_quant_results}

This section provides additional quantitative results for the tile-level procedural synthesis module, including full per-attribute results, visual
encoder ablations, metric definitions, and the synthetic-only versus
synthetic-plus-real training comparison.

\paragraph{Mixed-domain model evaluated on synthetic and real tiles.}
Tables~\ref{tab:app_window_synth_real} and \ref{tab:app_entrance_synth_real} evaluate the mixed-domain-trained component predictor on synthetic and real-world test tiles. They characterize the difference between the two evaluation domains.
The separate training-data comparison is reported in Appendix~\ref{app:synthetic_to_real_transfer}.

\begin{table}
\centering
\caption{Full attribute-level accuracy on window-like tiles.}
\label{tab:app_window_synth_real}
\small
\setlength{\tabcolsep}{6pt}
\begin{tabular}{llccc}
\toprule
Group & Attribute & Synthetic Acc.$\uparrow$ & Real-world Acc.$\uparrow$ & Gap$\downarrow$ \\
\midrule
\multirow{2}{*}{Arch structure}
& Arch style        & 0.96 & 0.93 & 0.03 \\
& Arch height class & 0.92 & 0.87 & 0.05 \\
\midrule
\multirow{3}{*}{Categorical}
& Arch        & 0.99 & 0.98 & 0.01 \\
& Top sill    & 0.96 & 0.92 & 0.04 \\
& Bottom sill & 0.95 & 0.93 & 0.02 \\
\midrule
\multirow{7}{*}{Frame attributes}
& Outer frame thickness & 0.89 & 0.86 & 0.03 \\
& Transom width         & 0.91 & 0.91 & 0.00 \\
& Mullion width         & 0.90 & 0.88 & 0.02 \\
& Row                   & 0.94 & 0.95 & -0.01 \\
& Row ratio             & 0.91 & 0.87 & 0.04 \\
& Columns per row       & 0.89 & 0.84 & 0.05 \\
& Column ratio per row  & 0.85 & 0.82 & 0.03 \\
\midrule
\multirow{4}{*}{Sill structure}
& Top sill height       & 0.93 & 0.88 & 0.05 \\
& Top sill protrusion   & 0.95 & 0.91 & 0.04 \\
& Bottom sill height    & 0.94 & 0.90 & 0.04 \\
& Bottom sill protrusion& 0.95 & 0.87 & 0.08 \\
\bottomrule
\end{tabular}
\end{table}

\begin{table}
\centering
\caption{Full attribute-level accuracy on entrance-like tiles.}
\label{tab:app_entrance_synth_real}
\small
\setlength{\tabcolsep}{5pt}
\resizebox{\linewidth}{!}{
\begin{tabular}{llccc}
\toprule
Group & Attribute & Synthetic Acc.$\uparrow$ & Real-world Acc.$\uparrow$ & Gap$\downarrow$ \\
\midrule
\multirow{3}{*}{Entrance frame}
& Frame style       & 0.97 & 0.93 & 0.04 \\
& Frame head height & 0.92 & 0.89 & 0.03 \\
& Frame side width  & 0.93 & 0.90 & 0.03 \\
\midrule
\multirow{5}{*}{Entrance structure}
& Door type             & 0.97 & 0.92 & 0.05 \\
& Door asymmetric ratio & 0.91 & 0.86 & 0.05 \\
& Door style            & 0.88 & 0.88 & 0.00 \\
& Grid rows             & 0.87 & 0.85 & 0.02 \\
& Grid columns          & 0.85 & 0.87 & -0.02 \\
\midrule
\multirow{3}{*}{Hardware}
& Has hardware  & 0.98 & 0.96 & 0.02 \\
& Hardware type & 0.97 & 0.96 & 0.01 \\
& Hardware side & 0.92 & 0.87 & 0.05 \\
\midrule
\multirow{3}{*}{Stairs}
& Has stairs   & 0.98 & 0.94 & 0.04 \\
& Stair height & 0.88 & 0.86 & 0.02 \\
& Stair count  & 0.90 & 0.86 & 0.04 \\
\bottomrule
\end{tabular}
}
\end{table}

\subsection{Full visual encoder ablation.}
\label{app:tile_ablation_results}

Table~\ref{tab:app_window_encoder_ablation} and Table~\ref{tab:app_entrance_encoder_ablation} report the full per-attribute accuracy of different visual encoders. We compare CLIP-only, DINOv3-only, and the combined CLIP+DINOv3 encoder. CLIP-only performs well on semantic presence attributes, while DINOv3 improves several structural attributes. The combined encoder achieves the most stable overall performance.

\begin{table}[t]
\centering
\caption{Full per-attribute encoder ablation on window-like tiles.}
\label{tab:app_window_encoder_ablation}
\small
\setlength{\tabcolsep}{4pt}
\resizebox{\linewidth}{!}{
\begin{tabular}{lccccccccc}
\toprule
Model & Has Arch & Arch Style & Arch Height & Has Sill & Sill Height & Sill Protrusion & Frame Thick. & Grid Num. & Grid Ratio \\
\midrule
CLIP-only   & 0.99 & 0.90 & 0.81 & 0.91 & 0.72 & 0.68 & 0.65 & 0.75 & 0.72 \\
DINOv3-only & 0.98 & 0.93 & 0.88 & 0.90 & 0.89 & 0.89 & 0.86 & 0.88 & 0.85 \\
CLIP+DINOv3 & \textbf{0.99} & \textbf{0.94} & \textbf{0.90} & \textbf{0.93} & \textbf{0.92} & \textbf{0.92} & \textbf{0.88} & \textbf{0.90} & \textbf{0.87} \\
\bottomrule
\end{tabular}
}
\end{table}

\begin{table}[t]
\centering
\caption{Full per-attribute encoder ablation on entrance-like tiles.}
\label{tab:app_entrance_encoder_ablation}
\small
\setlength{\tabcolsep}{4pt}
\resizebox{\linewidth}{!}{
\begin{tabular}{lcccccccccccc}
\toprule
Model & Has Stairs & Stair Height & Stair Count & Has Hardware & Hardware Type & Hardware Side & Frame Style & Frame Size & Door Type & Door Asym. Ratio & Door Style & Grid \\
\midrule
CLIP-only   & 0.95 & 0.74 & 0.71 & 0.95 & 0.94 & 0.80 & 0.93 & 0.82 & 0.86 & 0.78 & 0.86 & 0.77 \\
DINOv3-only & 0.93 & 0.83 & 0.85 & 0.93 & 0.91 & 0.84 & 0.89 & 0.88 & 0.90 & 0.85 & 0.87 & 0.84 \\
CLIP+DINOv3 & \textbf{0.96} & \textbf{0.87} & \textbf{0.88} & \textbf{0.96} & \textbf{0.96} & \textbf{0.88} & \textbf{0.95} & \textbf{0.90} & \textbf{0.94} & \textbf{0.88} & \textbf{0.88} & \textbf{0.86} \\
\bottomrule
\end{tabular}
}
\end{table}

\subsection{Attribute Discretization and Evaluation}
\label{app:attribute_metrics}

\paragraph{Normalized geometric presets.}
Geometric dimensions are normalized by the corresponding tile width or height and represented as attribute-specific preset categories. The recorded preprocessing procedure uses training-set quantiles to determine category boundaries. Attribute accuracy is evaluated by exact match against the annotated preset labels, while synthetic continuous-error evaluation compares preset-decoded values with the original CGA parameters.

\paragraph{Exact-match accuracy.}
Let $\mathcal V_j$ denote the evaluation tiles with valid ground-truth
labels for attribute $j$. Its exact-match accuracy is
\begin{equation}
\operatorname{Acc}_j=
\frac{1}{|\mathcal V_j|}
\sum_{i\in\mathcal V_j}
\mathbf{1}\!\left[\hat c_{ij}=c_{ij}\right],
\label{eq:app_attribute_accuracy}
\end{equation}
where $c_{ij}$ and $\hat c_{ij}$ are the annotated and predicted
attribute values or preset categories. The same exact-match definition
applies to categorical presence/type attributes and discrete counts.
Presence attributes, such as \textbf{Has stairs}, are evaluated on all relevant tiles with valid labels, whereas dependent attributes, such as \textbf{Stair height}, are evaluated only where the corresponding structure is present in the ground truth and their labels are valid.

\paragraph{Continuous error on synthetic test tiles.}
For synthetic samples, the original continuous CGA parameters are
available. We therefore also compare the normalized continuous target
$\tilde v_{ij}$ with the representative value $r_j(\hat c_{ij})$ of the
predicted preset. For the synthetic samples $\mathcal S_j$ with a valid
continuous target for attribute $j$,
\begin{equation}
\operatorname{MAE}_j=
\frac{1}{|\mathcal S_j|}
\sum_{i\in\mathcal S_j}
\left|r_j(\hat c_{ij})-\tilde v_{ij}\right|.
\label{eq:app_synthetic_mae}
\end{equation}
This measures the continuous error of the preset-decoded prediction,
rather than the output of a continuous regression model. The error is
dimensionless in normalized tile coordinates, not meters.
Table~\ref{tab:app_synthetic_continuous_error} reports the results.

\begin{table}[htbp]
\centering
\caption{Synthetic-test preset accuracy and normalized continuous error.
The representative value of the predicted preset is compared with
the normalized continuous CGA parameter.}
\label{tab:app_synthetic_continuous_error}
\small
\begin{tabular}{lcc}
\toprule
Attribute group & Exact Acc.$\uparrow$ & MAE$\downarrow$ \\
\midrule
Window frame geometry & 0.900 & 0.026 \\
Window grid ratios    & 0.880 & 0.031 \\
Window sill geometry  & 0.943 & 0.016 \\
Entrance geometry     & 0.910 & 0.024 \\
\bottomrule
\end{tabular}
\end{table}

\subsection{Synthetic-Only versus Synthetic-Plus-Real Training}
\label{app:synthetic_to_real_transfer}

We compare a component predictor trained with synthetic data only against
one trained with the same synthetic data augmented by the real training
partition. Both are evaluated on the same held-out set of 107 real tiles
from 36 building groups (Appendix~\ref{app:real_tile_split}). Only the
synthetic-plus-real setting uses the 100 real training crops.
This experiment tests the effect of adding real-domain supervision,
separately from evaluating a mixed-domain-trained model on two test domains.

\begin{table}[htbp]
\centering
\caption{Training-data comparison on the same held-out real test partition. W and E denote window-like and entrance-like components. Overall macro is the arithmetic mean of the five reported grouped metrics.}
\label{tab:app_synthetic_real_training}
\small
\setlength{\tabcolsep}{4pt}
\resizebox{\linewidth}{!}{%
\begin{tabular}{lcccccc}
\toprule
Training data & Overall macro$\uparrow$ & W-Arch$\uparrow$ &
W-Row/Col$\uparrow$ & W-Sill$\uparrow$ & E-Stair$\uparrow$ & E-Handle$\uparrow$ \\
\midrule
Synthetic only   & 0.868 & 0.91 & 0.82 & 0.83 & 0.89 & 0.89 \\
Synthetic + real & 0.940 & 0.98 & 0.89 & 0.92 & 0.95 & 0.96 \\
\bottomrule
\end{tabular}%
}
\end{table}

The representative attribute scores improve with real training crops:
W-Arch increases from $0.91$ to $0.98$, W-Row/Col from $0.82$ to $0.89$,
W-Sill from $0.83$ to $0.92$, E-Stair from $0.89$ to $0.95$, and E-Handle
from $0.89$ to $0.96$. These results support synthetic-to-real transfer
and the benefit of real-domain calibration on the evaluated
building-disjoint test partition.

\section{Additional Qualitative Results}
\label{app:additional_qualitative_results}

\subsection{Additional Global Structure Layout Recovery Results}
\label{app:additional_macro_results}

\begin{figure}[t]
    \centering
    \includegraphics[width=\linewidth]{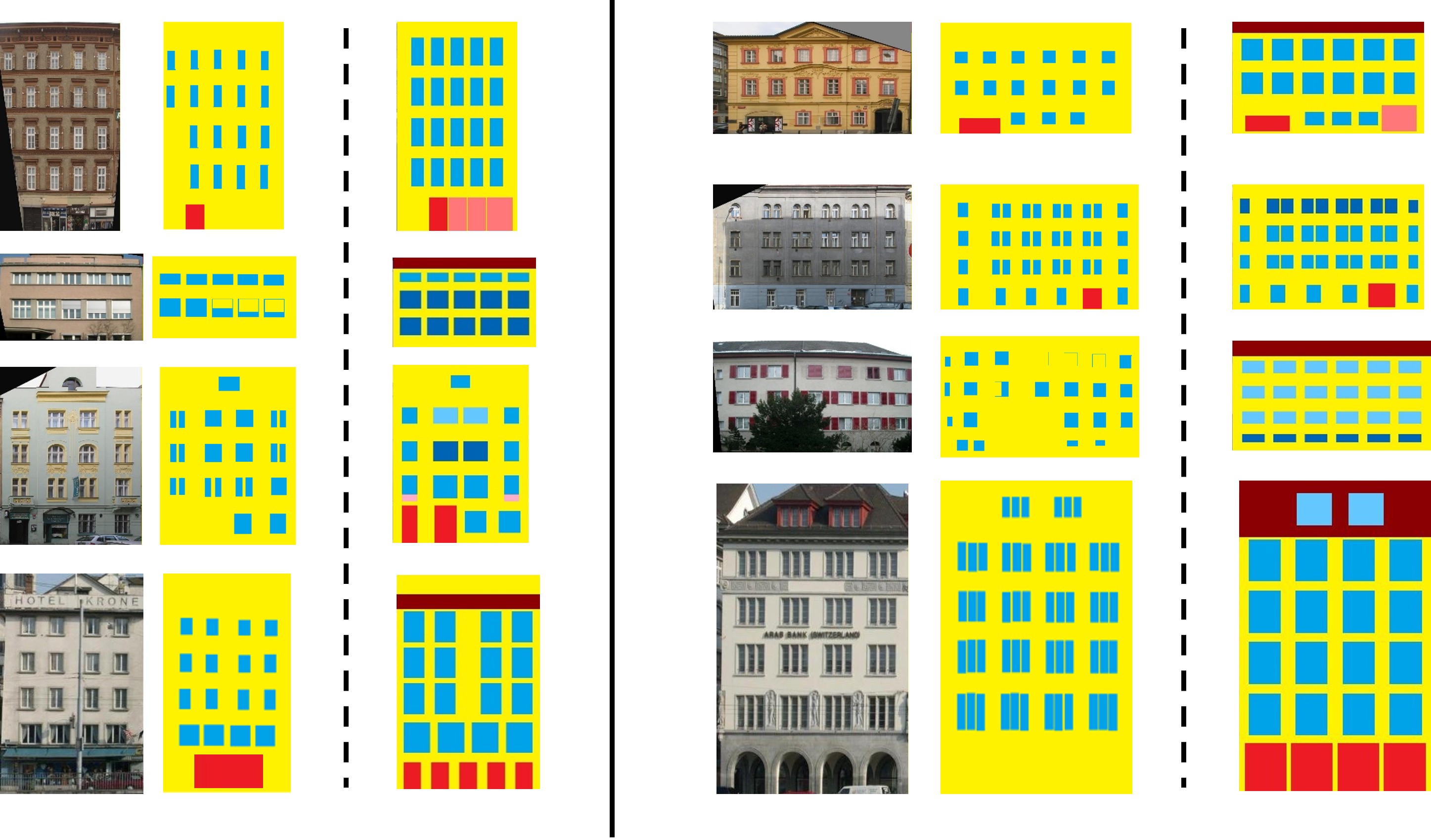}
    \caption{\textbf{Additional global structure layout recovery results on regular CMP facades.}
    Each example shows the input facade image, the remapped structural reference, and the recovered global structure layout.}
    \label{fig:app_cmp_results}
\end{figure}

\paragraph{Regular CMP facades.}
Fig.~\ref{fig:app_cmp_results} presents additional global structure layout recovery results on regular facade examples from CMP. For these cases, the provided annotations are remapped into our visualization scheme and used as structural references for layout-level comparison. These additional examples further demonstrate that the proposed global structure recovery stage can infer stable floor-wise and bay-wise facade organization from RGB images.

\paragraph{Challenging facades.}
Fig.~\ref{fig:app_challenging_results} shows additional results on challenging facade images from eTRIMS and in-the-wild examples. These cases include perspective distortion, partial occlusion, nonuniform spacing, and incomplete visual evidence. The results provide further qualitative evidence that the reflection-based layout recovery stage can recover structurally meaningful facade layouts under difficult visual conditions.

\begin{figure}[t]
    \centering
    \includegraphics[width=\linewidth]{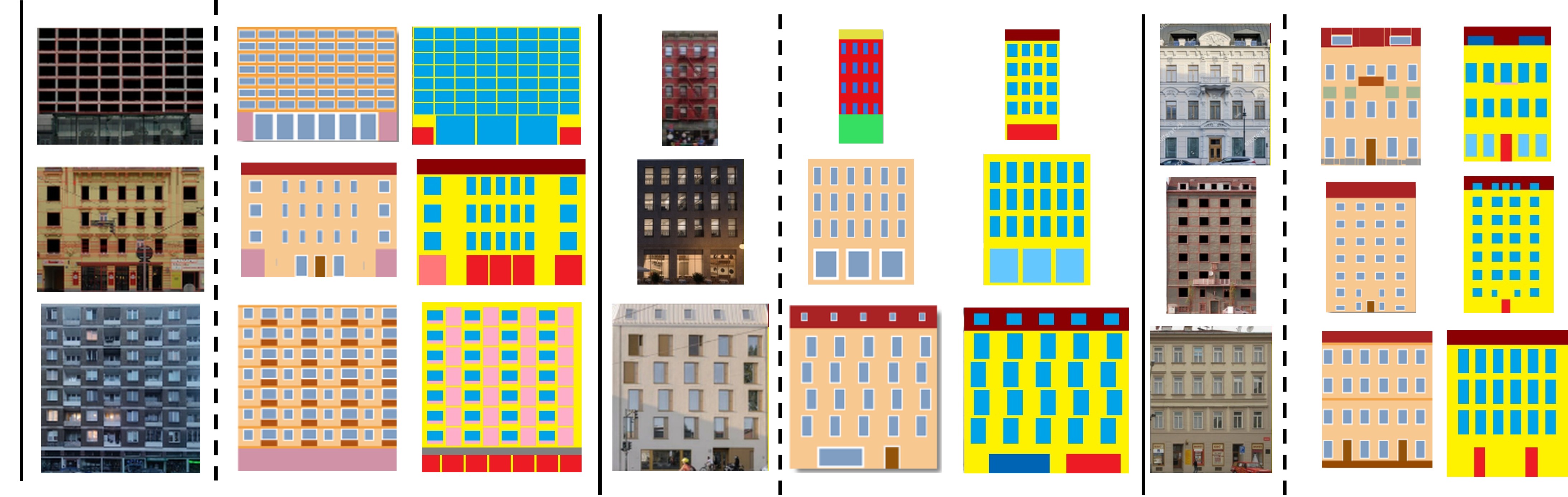}
    \caption{\textbf{Additional qualitative comparison with FaçAID.}
    Each example shows the input facade image, the corresponding FaçAID result, and our recovered layout.}
    \label{fig:app_facaid_results}
\end{figure}

\paragraph{Additional comparison with FaçAID.}
Fig.~\ref{fig:app_facaid_results} provides additional qualitative comparisons with representative examples reported in FaçAID. Since FaçAID assumes segmented facade input and no unified executable evaluation protocol is available, these comparisons are intended as qualitative layout-level references. Each example compares the input image, the FaçAID result, and our recovered layout.

\subsection{Additional Comparison Results}
\label{app:direct_generation_examples}

\begin{figure}[t]
    \centering
    \includegraphics[width=\linewidth]{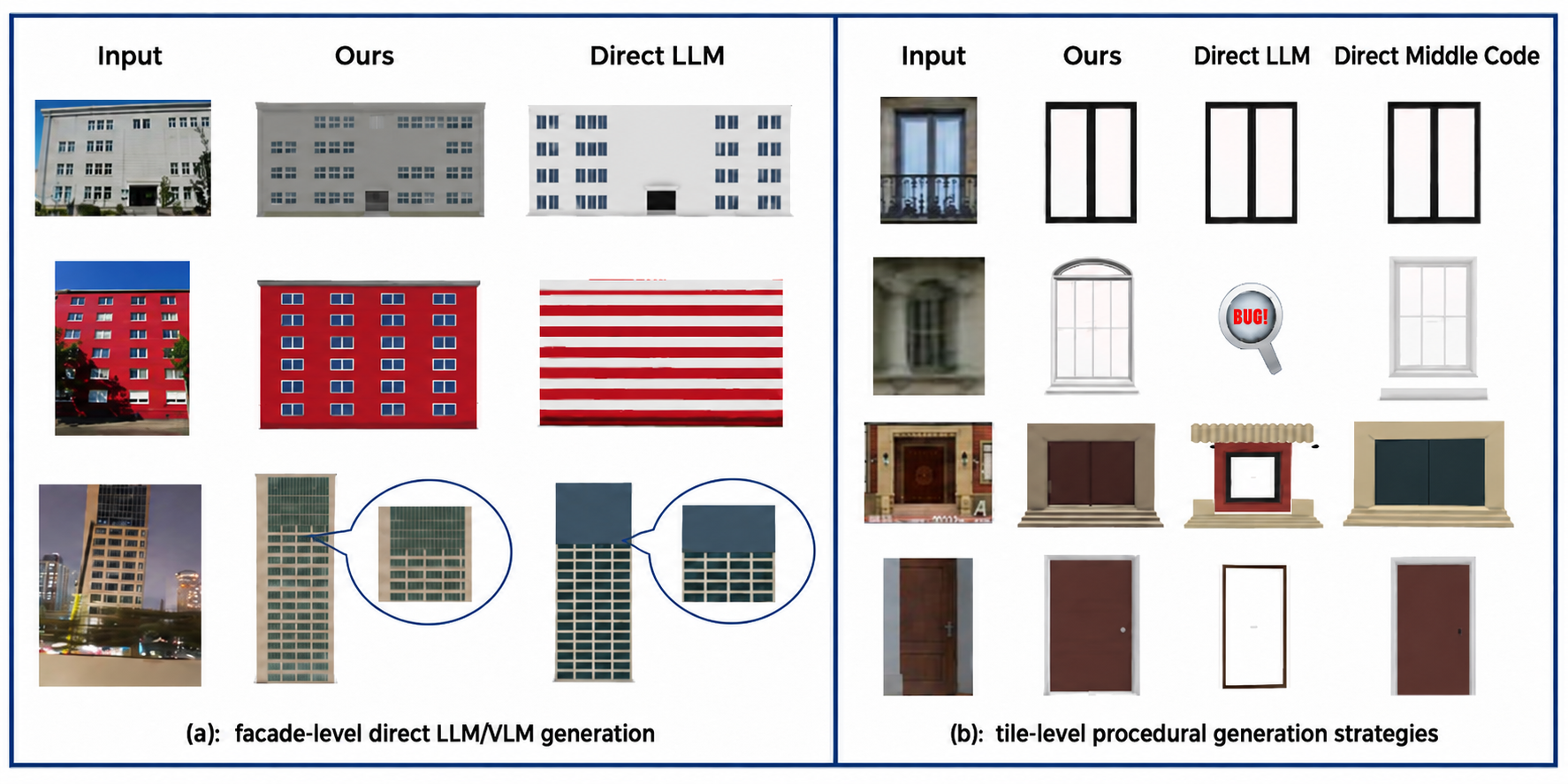}
    \caption{\textbf{Additional qualitative comparisons between our progressive reasoning workflow and direct LLM/VLM outputs.}
(a) Facade-level comparison between our full framework and direct
end-to-end CGA generation by a general LLM/VLM~\citep{gemini15}.
From left to right are the input facade image, our result, and the direct LLM/VLM result.
(b) Tile-level comparison among our fine-tuned tile-level predictor, direct low-level code generation, and direct middle-code prediction. From left to right are the input tile crop, our result, the direct low-level code generation result, and the direct middle-code result.}
    \label{fig:app_direct_generation_examples}
\end{figure}

Our method adopts a progressive reasoning workflow that infers
appropriate facade structures through a dynamic reflection loop.
To further illustrate its advantages, Fig.~\ref{fig:app_direct_generation_examples}
provides additional qualitative comparisons against direct LLM/VLM baselines. The left panel compares facade-level direct end-to-end generation with our full framework, while the right panel compares tile-level procedural generation strategies, including our fine-tuned predictor, direct low-level code generation, and direct middle-code prediction. These examples complement the quantitative comparisons in the main paper.

\subsection{Additional Editable Tile-Level Results}
\label{app:tile_editable_results}

To further illustrate the interpretability and editability of the proposed component parameterization, we provide additional tile-level examples for representative window-like and entrance-like components. These examples show that the recovered structured tile descriptions can be translated into editable procedural assets while preserving the main local architectural properties.

\begin{figure}[t]
    \centering
    \includegraphics[width=0.9\linewidth]{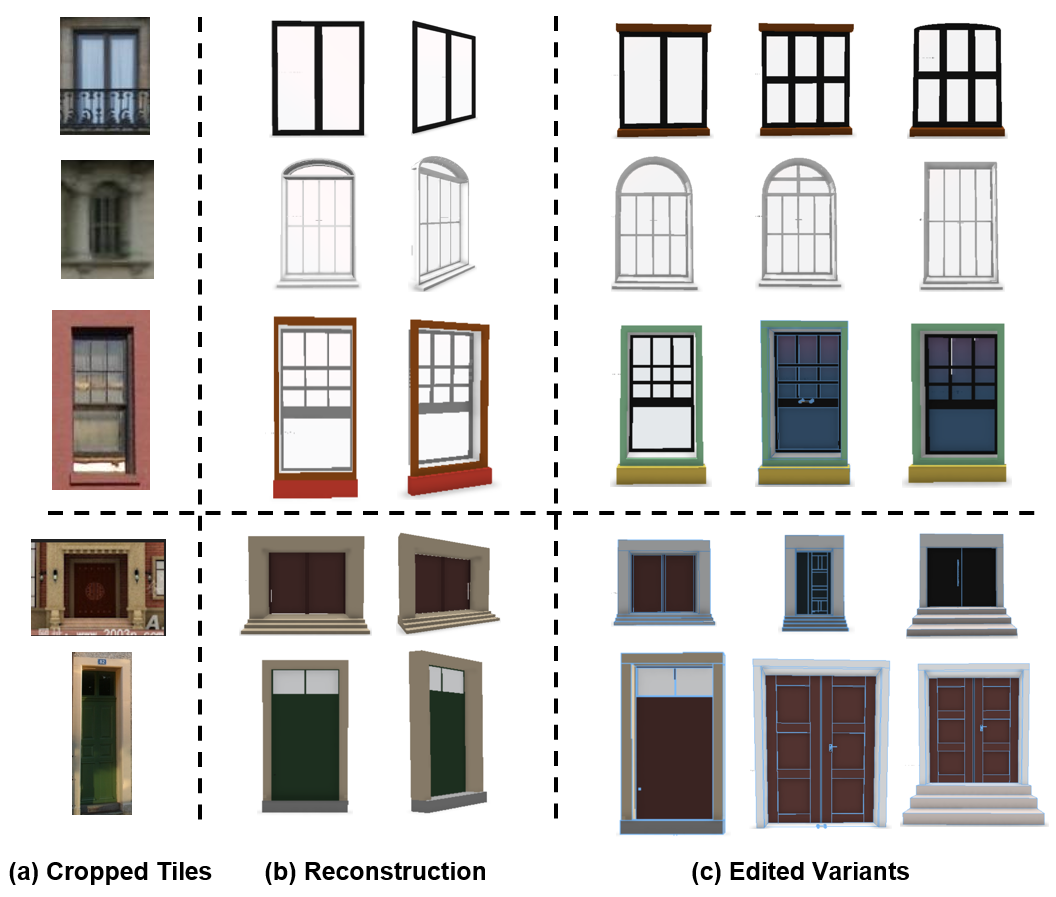}
    \caption{\textbf{Editable tile-level results.}
    Top: representative window-like tiles. Bottom: representative entrance-like tiles. For each example, we show the input crop and the corresponding procedural reconstruction generated from the predicted tile-level representation.}
    \label{fig:app_tile_editable}
\end{figure}

\paragraph{Window-like tiles.}
Fig.~\ref{fig:app_tile_editable} shows additional editable reconstruction examples for window-like tiles. These examples cover variations in arch presence, grid subdivision, sill configuration, and frame structure.

\paragraph{Entrance-like tiles.}
Fig.~\ref{fig:app_tile_editable} presents additional editable reconstruction examples for entrance-like tiles. The examples include different doorway layouts, stair configurations, frame structures, and hardware-related details.

\section{User Study Protocol}
\label{app:user_study}

We conduct a user preference study to compare our recovered facade layouts
with Fa\c{c}AID. The study is designed as a supplementary perceptual evaluation
rather than a replacement for quantitative structural evaluation.

\subsection{Participants}

The study involved 15 participants, including 5 participants with
architecture-related backgrounds and 10 participants without architecture
training. The participants consisted of 8 males and 7 females.

\subsection{Evaluation Protocol}

Each participant evaluated all 12 paired facade examples. For each trial,
participants were shown:
(1) the input facade image and
(2) two anonymized layout reconstruction results.

The two methods were randomly assigned to the left or right position for each
trial, and method identities were hidden. Participants were asked to select
the layout that better preserves the dominant facade structure.

The instruction provided to participants was:

\begin{quote}
Given the input facade and two candidate layout reconstructions, select the
result that better preserves the dominant facade structure, including floor
organization, window and entrance placement, major facade regions, and
repetition patterns. Ignore color and rendering style. Select ``Tie'' if the
two results are comparably faithful or if no reliable preference can be made.
\end{quote}

\subsection{Preference Results}

The study produced 106 votes for our method, 54 for Fa\c{c}AID, and 20 ties, corresponding to 58.9\%, 30.0\%, and 11.1\%, respectively. Among the 160 non-tie judgments, our method received 66.3\% of the votes. Under an independent-vote binomial model, the non-tie preference proportion has a nominal exact 95\% confidence interval of $[58.4\%,73.5\%]$ (Clopper--Pearson), with a two-sided $p=4.80\times10^{-5}$ against $p_0=0.5$. These unadjusted statistics do not account for participant- or example-level dependence and are reported only as exploratory evidence. We interpret these descriptive results as supplementary perceptual evidence on the 12 evaluated examples, rather than a comprehensive benchmark against Fa\c{c}AID.

\section{TRC Evaluation Protocol}
\label{app:trc_protocol}

Tree Repair Cost (TRC) measures the expert-validated structural repair cost under a fixed editing protocol. It is designed to evaluate how much manual structural correction is required for a predicted layout tree. TRC is evaluated by expert annotators who are separate from the participants in the Fa\c{c}AID preference study.

For each evaluation example, we manually construct one valid reference layout
and render it into a schematic representation. The reference schematic serves
as a visual structural target; the predicted tree is not required to match the
reference tree in split order, grouping strategy, or serialization.

We develop an annotation interface containing:
(1) the input facade image,
(2) the rendered reference schematic,
(3) the predicted layout tree and its rendered schematic,
and
(4) editing operations for modifying the predicted tree.

The annotator edits only the predicted layout tree until its rendered schematic
expresses the same dominant facade organization as the accepted reference.
If the predicted tree already produces an acceptable reconstruction, its TRC
is zero.

One atomic edit is defined as:

\begin{itemize}
    \item inserting one structural node;
    \item deleting one structural node;
    \item modifying one existing node, including operator type, split axis,
    repeat count, span, or related attributes.
\end{itemize}

When several attributes of the same node are corrected together, they are counted as one node modification.

All annotators receive the same operation definitions and representative examples. Each sample is independently evaluated, and the lowest validated repair count is retained as its TRC.

Therefore, TRC should be interpreted as the lowest expert-validated repair cost under the fixed protocol, rather than a guaranteed global minimum over all possible procedural decompositions. We do not report a formal inter-annotator agreement statistic and interpret TRC as an expert-assisted structural diagnostic.




    

\clearpage

\section*{NeurIPS Paper Checklist}

\begin{enumerate}

\item {\bf Claims}
    \item[] Question: Do the main claims made in the abstract and introduction accurately reflect the paper's contributions and scope?
    \item[] Answer: \answerYes{} 
    \item[] Justification:  The abstract and introduction state that the method focuses on dominant visible facade reconstruction and editable building asset instantiation, rather than full reconstruction of unseen geometry. The method, experiments, and conclusion are aligned with this scope.

\item {\bf Limitations}
    \item[] Question: Does the paper discuss the limitations of the work performed by the authors?
    \item[] Answer: \answerYes{}
    \item[] Justification: The paper discusses the scope and assumptions of the method, including its focus on the dominant visible facade, the orthogonal assumption, and the use of a simple mass model for editable building asset instantiation. The conclusion and method sections also clarify that the recovered visible-facade program is the primary output, rather than a full reconstruction of unseen building geometry.

    \item {\bf Theory assumptions and proofs}
        \item[] Question: For each theoretical result, does the paper provide the full set of assumptions and a complete (and correct) proof?
        \item[] Answer: \answerNA{}
        \item[] Justification: The paper does not present theoretical results, theorems, or formal proofs. The contribution is an empirical and methodological framework for visual-to-executable procedural reconstruction.
    
        \item {\bf Experimental result reproducibility}
        \item[] Question: Does the paper fully disclose all the information needed to reproduce the main experimental results of the paper to the extent that it affects the main claims and/or conclusions of the paper (regardless of whether the code and data are provided or not)?
        \item[] Answer: \answerYes{}
        \item[] Justification: The paper describes the model components, the global structure recovery and component parameterization pipeline, the deterministic CGA compilation process, the evaluation metrics, and the datasets used for macro-structure recovery and tile-level evaluation. Additional training and implementation details, representation schemas, and extended quantitative results are provided in the appendix.

\item {\bf Open access to data and code}
    \item[] Question: Does the paper provide open access to the data and code, with sufficient instructions to faithfully reproduce the main experimental results, as described in supplemental material?
    \item[] Answer: \answerNo{}
    \item[] Justification: The paper does not release code, model checkpoints, or newly constructed training data at submission time. However, the main method, representation schemas, deterministic compilation process, evaluation metrics, and training/implementation details are described in the main paper and appendix to support reproducibility.

\item {\bf Experimental setting/details}
    \item[] Question: Does the paper specify all the training and test details (e.g., data splits, hyperparameters, how they were chosen, type of optimizer) necessary to understand the results?
    \item[] Answer: \answerYes{}
    \item[] Justification: The main paper describes the evaluation settings, datasets, baselines, and metrics used for global structure recovery, component parameterization, direct generation comparison, and ablation studies. Additional training and implementation details, including data construction and model usage, are provided in the appendix.

\item {\bf Experiment statistical significance}
    \item[] Question: Does the paper report error bars suitably and correctly defined or other appropriate information about the statistical significance of the experiments?
    \item[] Answer: \answerNo{}
    \item[] Justification: Appendix~\ref{app:user_study} reports an exploratory vote-level binomial analysis with its independence assumption and limitations. We do not report participant- or example-adjusted uncertainty estimates, or repeated-run uncertainty estimates for the other quantitative experiments.

\item {\bf Experiments compute resources}
    \item[] Question: For each experiment, does the paper provide sufficient information on the computer resources (type of compute workers, memory, time of execution) needed to reproduce the experiments?
    \item[] Answer: \answerNo{}
    \item[] Justification: The paper reports the model architecture, training settings, evaluation protocol, and deterministic compilation pipeline. Detailed hardware specifications, runtime measurements, and total computational costs are not reported.
    
\item {\bf Code of ethics}
    \item[] Question: Does the research conducted in the paper conform, in every respect, with the NeurIPS Code of Ethics \url{https://neurips.cc/public/EthicsGuidelines}?
    \item[] Answer: \answerYes{}
    \item[] Justification: The research uses architectural images and procedural modeling techniques and does not involve sensitive personal data, biometric identification, surveillance, or high-risk decision-making. We preserve submission anonymity and follow the NeurIPS Code of Ethics throughout the study.

\item {\bf Broader impacts}
    \item[] Question: Does the paper discuss both potential positive societal impacts and negative societal impacts of the work performed?
    \item[] Answer: \answerYes{}
    \item[] Justification: The paper discusses the intended use of the method for editable architectural and urban modeling. Potential positive impacts include reducing the cost of procedural building asset creation, while potential negative impacts include over-reliance on automatically reconstructed building geometry in applications that require accurate measurements or safety-critical decisions.
    
\item {\bf Safeguards}
    \item[] Question: Does the paper describe safeguards that have been put in place for responsible release of data or models that have a high risk for misuse (e.g., pre-trained language models, image generators, or scraped datasets)?
    \item[] Answer: \answerNo{}
    \item[] Justification: The paper does not release a high-risk pretrained model, image generator, or scraped dataset. The method focuses on procedural reconstruction of building facades and is not intended for surveillance, biometric identification, or safety-critical decision-making.

\item {\bf Licenses for existing assets}
    \item[] Question: Are the creators or original owners of assets (e.g., code, data, models), used in the paper, properly credited and are the license and terms of use explicitly mentioned and properly respected?
    \item[] Answer: \answerYes{}
    \item[] Justification: The paper cites the existing datasets, models, and related methods used for evaluation and comparison, including facade datasets, prior procedural modeling methods, and LLM/VLM baselines. We use these assets for research evaluation and follow their intended academic use.

\item {\bf New assets}
    \item[] Question: Are new assets introduced in the paper well documented and is the documentation provided alongside the assets?
    \item[] Answer: \answerNo{}
    \item[] Justification: The paper constructs synthetic and real training examples for layout recovery and component parameterization, but these assets are not released at submission time. The paper and appendix document the data construction procedure, representation schemas, and annotation format to support understanding and future reproducibility.

\item {\bf Crowdsourcing and research with human subjects}
    \item[] Question: For crowdsourcing experiments and research with human subjects, does the paper include the full text of instructions given to participants and screenshots, if applicable, as well as details about compensation (if any)?
    \item[] Answer: \answerNo{}
    \item[] Justification: Appendix~\ref{app:user_study} provides participant information, the evaluation protocol, and the full instructions given to participants. Interface screenshots and compensation details are not reported. The study is used as supplementary perceptual evidence.

\item {\bf Institutional review board (IRB) approvals or equivalent for research with human subjects}
    \item[] Question: Does the paper describe potential risks incurred by study participants, whether such risks were disclosed to the subjects, and whether Institutional Review Board (IRB) approvals (or an equivalent approval/review based on the requirements of your country or institution) were obtained?
    \item[] Answer: \answerNo{}
    \item[] Justification: The paper includes a small user preference study involving anonymized visual comparisons, but it does not report IRB approval or equivalent review. The study does not collect sensitive personal information and is not used for high-risk decision-making.

\item {\bf Declaration of LLM usage}
    \item[] Question: Does the paper describe the usage of LLMs if it is an important, original, or non-standard component of the core methods in this research? Note that if the LLM is used only for writing, editing, or formatting purposes and does \emph{not} impact the core methodology, scientific rigor, or originality of the research, declaration is not required.
    \item[] Answer: \answerYes{}
    \item[] Justification: The core method uses a fine-tuned vision-language model for global structure layout induction and visual-symbolic reflection, and also compares against LLM/VLM direct generation baselines. These uses are described in the method, experiments, and appendix.

\end{enumerate}

\end{document}